\documentclass[conference]{IEEEtran}
\IEEEoverridecommandlockouts
\usepackage{cite}
\usepackage{amsmath,amssymb,amsfonts}
\usepackage{algorithmic}
\usepackage{graphicx}
\usepackage{textcomp}
\usepackage{xcolor}
\usepackage{epsfig,tabularx,booktabs,comment,multirow,algorithm,enumitem}
\def\BibTeX{{\rm B\kern-.05em{\sc i\kern-.025em b}\kern-.08em
    T\kern-.1667em\lower.7ex\hbox{E}\kern-.125emX}}
\begin{document}

\title{NID-SLAM: Neural Implicit Representation-based RGB-D SLAM In Dynamic Environments\\
}

\author{\IEEEauthorblockN{1\textsuperscript{st} Ziheng Xu}
	\IEEEauthorblockA{\textit{School of Computer Science} \\
		\textit{Beihang University}\\
		Beijing, China \\
		xuziheng0310@buaa.edu.cn}
	\and
	\IEEEauthorblockN{2\textsuperscript{nd} Jianwei Niu}
	\IEEEauthorblockA{\textit{School of Computer Science} \\
		\textit{Beihang University}\\
		Beijing, China \\
		niujianwei@buaa.edu.cn}
	\and
	\IEEEauthorblockN{3\textsuperscript{rd} Qingfeng Li}
	\IEEEauthorblockA{\textit{School of Computer Science} \\
		\textit{Beihang University}\\
		Beijing, China \\
		liqingfeng@buaa.edu.cn}
	\and
	\IEEEauthorblockN{4\textsuperscript{th} Tao Ren\IEEEauthorrefmark{2}}
	\IEEEauthorblockA{\textit{State Key Laboratory of Intelligent Game} \\
		\textit{Institute of Software Chinese Academy of Sciences}\\
		Beijing, China \\
		rentao22@iscas.ac.cn}
	\and
	\IEEEauthorblockN{5\textsuperscript{th} Chen Chen}
	\IEEEauthorblockA{\textit{Big Data and Industrial Intelligent Networking Laboratory}\\
		\textit{Hangzhou Innovation Institute of Beihang University}\\
		Hangzhou, China \\
		chenx2ovo@buaa.edu.cn}
	\thanks{\IEEEauthorrefmark{2} Corresponding author.}
}
\maketitle

\begin{abstract}
Neural implicit representations have been explored to enhance visual SLAM algorithms, especially in providing high-fidelity dense map. Existing methods operate robustly in static scenes but struggle with the disruption caused by moving objects. In this paper we present NID-SLAM, which significantly improves the performance of neural SLAM in dynamic environments. We propose a new approach to enhance inaccurate regions in semantic masks, particularly in marginal areas. Utilizing the geometric information present in depth images, this method enables accurate removal of dynamic objects, thereby reducing the probability of camera drift. Additionally, we introduce a keyframe selection strategy for dynamic scenes, which enhances camera tracking robustness against large-scale objects and improves the efficiency of mapping. Experiments on publicly available RGB-D datasets demonstrate that our method outperforms competitive neural SLAM approaches in tracking accuracy and mapping quality in dynamic environments.
\end{abstract}


\begin{figure}[t]
	\centering
	\begin{minipage}[b]{.32\linewidth}
		\centering
		{\epsfig{figure=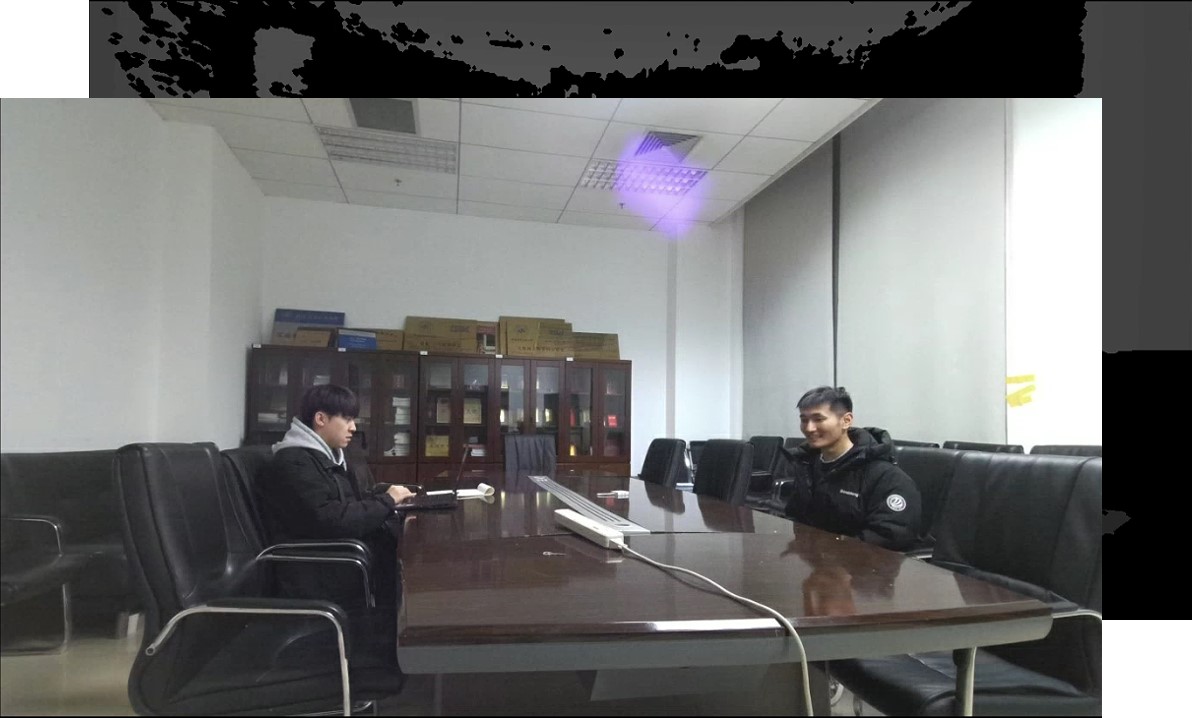,width=2.7cm}}
	\end{minipage}
	\begin{minipage}[b]{0.32\linewidth}
		\centering
		{\epsfig{figure=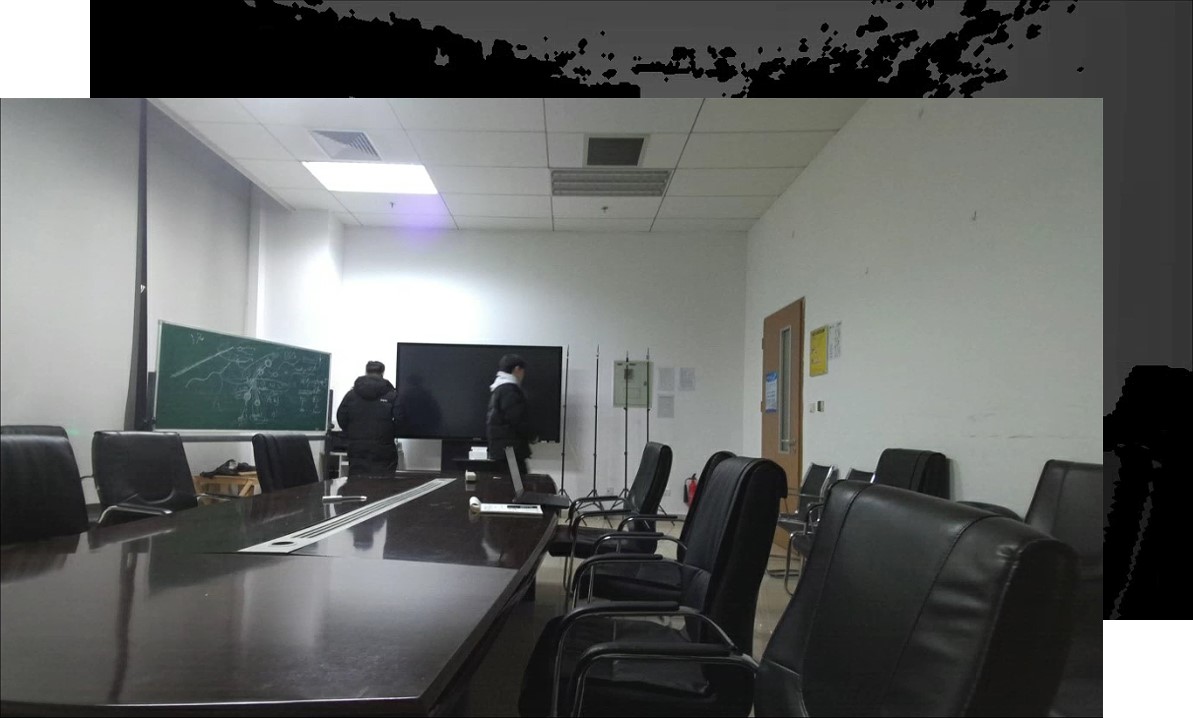,width=2.7cm}}
	\end{minipage}
	\begin{minipage}[b]{0.32\linewidth}
		\centering
		{\epsfig{figure=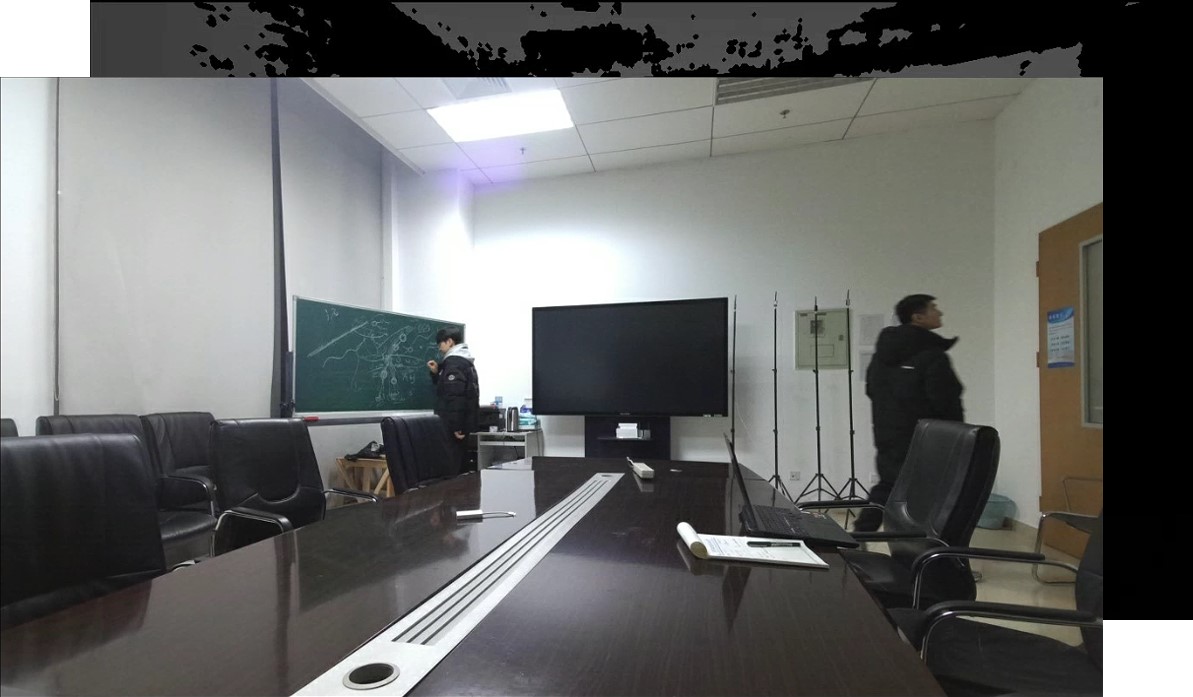,width=2.7cm}}
	\end{minipage}
	\medskip\centering{(a) Input RGB-D frames with dynamic objects.}
	\begin{minipage}[b]{1.0\linewidth}
		\centering
		{\epsfig{figure=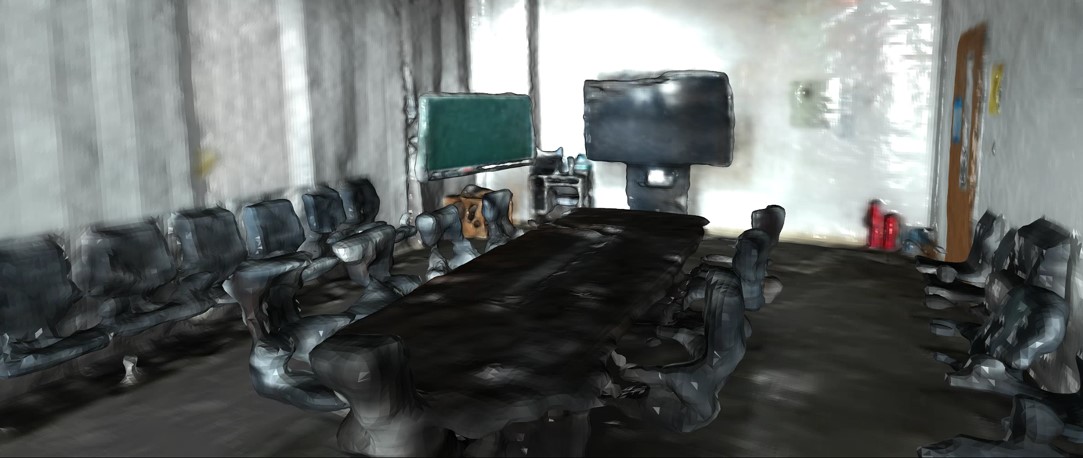,width=8.5cm}}
	\end{minipage}
	\centering{(b) Output map of the scene after removing dynamic objects.}
	\caption{The 3D reconstruction results of NID-SLAM on our self-captured large dynamic scene.}
	\label{fig:Figure1}
\end{figure} 

\section{Introduction}
Visual simultaneous localization and mapping (SLAM) plays a crucial role in various applications such as robot navigation, augmented reality (AR), and virtual reality (VR). Visual SLAM algorithms utilize data collected by sensors, e.g., 
monocular, stereo and RGB-D cameras, to estimate the camera pose in previously unknown environments and incrementally build maps of the surrounding scenes \cite{1}. Among diverse visual sensors, RGB-D cameras concurrently record color and depth data, offering a more efficacious and precise basis for the acquisition of three-dimensional environmental information. This capability enhances the 3D reconstruction performances of most SLAM algorithms.

Recent methods \cite{9} have introduced neural implicit representation into SLAM. The most notable example, neural radiance fields (NeRF) \cite{4}, encodes scene color and volume density into the weights of a neural network, directly learning high-frequency scene details from the data, significantly enhancing mapping smoothness and coherence. Combined with rendering methods based on volumetric representation \cite{13}, NeRF, through training, can re-synthesize input images and generalize to adjacent, unseen viewpoints. However, most neural SLAM algorithms are based on the assumption of a static environment, with some capable of handling small dynamic objects in synthetic scenes \cite{10}. In real-world dynamic scenarios, these algorithms could experience a significant performance-degradation in terms of dense reconstruction and camera tracking accuracy. This could be largely caused by the existence of dynamic objects in the scene that induce incorrect data associations, thereby severely disrupting the pose estimation during the tracking process. Additionally, information about dynamic objects is often incorporated into the map, hindering its long-term applicability.

Semantic information has been introduced in many studies to address the challenges faced by visual SLAM algorithms in dynamic scenes \cite{2, 3}. The main idea is to integrate semantic information with geometric constraints to eliminate dynamic objects from the scene. However, on the one hand, due to the reduction of static information in the scene, the texture and coherence of the map  in these algorithms are poor. On the other hand, due to the absence of reasonable geometric prediction capabilities for unobserved areas, these algorithms often suffer from the non-negligible holes contained in the restored backgrounds.

To address the issue, we propose Neural Implicit Dynamic SLAM (NID-SLAM). We integrate accuracy-improved depth information with semantic segmentation to detect and remove dynamic objects, and project the static map into the current frame to inpaint backgrounds occluded by these objects. Overall, we make the following contributions:
\begin{itemize}[topsep=0pt, partopsep=0pt, itemsep=0pt, parsep=0pt]
	\item We present a neural implicit representation-based approach capable of achieving robust camera tracking and predictive, high-quality mapping in real-world dynamic environments.
	\item We propose a depth-guided semantic mask enhancement method to largely eliminate inaccuracies along edge regions.
	\item We introduce a dynamic scene-oriented keyframe selection strategy, which reduces the impact of unreliable frames, thereby improving tracking accuracy.
\end{itemize}

\section{RELATED WORK}

The masks, obtained from semantic segmentation networks, effectively label dynamic objects in images, serving as stable and reliable prior constraints to guide dynamic SLAM systems. 
Methods such as \cite{2, 3, 8} typically integrate semantic and geometric constraints, yet lack attention to depth information, leading to less precise priors and may include noise blocks. SaD-SLAM \cite{7} adopts semantic-depth-based movable object tracking, offering enhanced accuracy, while Blitz-SLAM \cite{12} revises the object mask using depth information. Contrasting these methods, our NID-SLAM dynamically detects and eliminates areas with inaccurate depth information, thus specifically enhancing the edge areas of semantic masks that are prone to errors.

Neural implicit represetations \cite{4, 18, 19}, which encode the 3D geometry and appearance of scenes into the weights of neural networks, have gained popularity due to their compact and precise results. When applied to visual SLAM, this technique enables the construction of environmental maps and the estimation of camera poses. 
Previous methods \cite{9, 10, 20, 21, 11, 22} have demonstrated the feasibility of using neural networks to model the color and geometric information of static scenes; however, they have not fully exploited the potential of neural implicit representations in dynamic environments. Instead of the approach used in \cite{10}, which resolves small dynamic objects in synthetic scenes by discarding outliers in losses, we employ a comprehensive removal method to address large dynamic objects in real-world scenes.

\begin{figure*}[t]
	\centering
	\includegraphics[width=\textwidth]{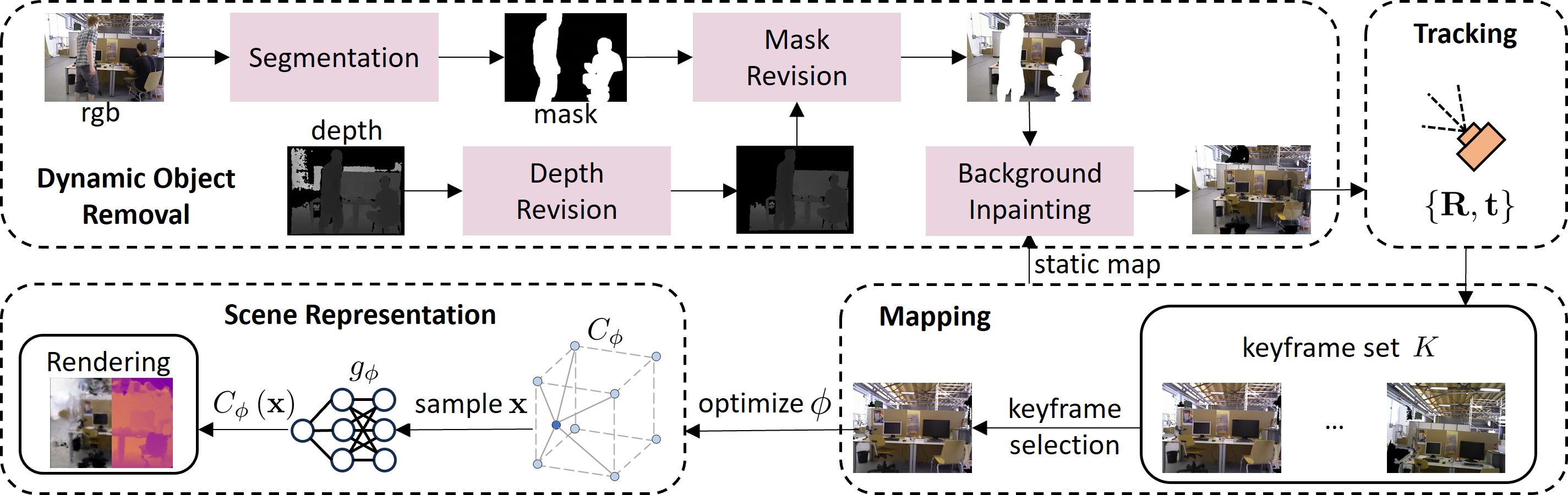}
	\caption{\textbf{System overview.} 1) Dynamic object removal: by employing semantic segmentation and mask revision, dynamic objects within RGB-D images are precisely eliminated, followed by a thorough restoration of the occluded backgrounds. 2) Tracking: camera poses $\{\mathbf{R}, \mathbf{t}\}$ are optimized by minimizing losses. 3) Mapping: a mask-guided strategy is employed to select keyframes for the optimization of feature grids scene representation. 4) Scene representation: efficient rendering of predicted color and depth values is achieved through surface-focused point sampling.}
	\label{fig:Figure2}
\end{figure*} 

\section{Methods}
Figure~\ref{fig:Figure2} illustrates the overall framework of NID-SLAM. Given an RGB-D image stream as input, we first remove dynamic objects using a specialized dynamic processing procedure. Subsequently, we accomplish tracking and mapping by jointly optimizing camera poses and neural scene representations. Utilizing semantic priors and depth information, dynamic objects are eliminated, and occluded backgrounds are restored through the static map. In each iteration of mapping, keyframes are selected to optimize scene representations and camera poses. Rendering is performed by sampling viewing rays and integrating predicted values at various points along these rays.

\subsection{Dynamic object removal}
\noindent\textbf{Depth revision.}
Due to the limitations of depth cameras, the accuracy of depth estimation decreases as the distance between the object and the camera increases. Depth information with significant errors can result in incorrect data associations, disrupting the stability of camera tracking. In highly dynamic environments, these inaccuracies become more pronounced, increasing the probability of camera drift. Additionally, due to errors in depth information, the constructed map can exhibit layering phenomena, where image blocks that should be at the same depth are represented at different depths on the map. Therefore, we detect and remove inaccurate depth information. Specifically, we compute the image gradients of the depth map, utilizing these gradients as indicators to assess the accuracy of depth information. When the horizontal or vertical gradient of an image surpasses a predefined threshold $\tau_1$, indicating significant depth variation, we set the depth of the subsequent pixel point in the gradient direction to zero to mitigate depth errors.

\noindent\textbf{Depth-based semantic segmentation.}
To detect dynamic objects, we employ a bounding box-based network for semantic segmentation of the input image, using the YOLO algorithm in our experiments. The network takes RGB raw images as input and outputs binary masks of potential dynamic or movable objects in the image. The semantic masks exhibit two primary shortcomings. Firstly, they may not fully cover dynamic objects, potentially incorporating other objects from the environment. Secondly, the masks are prone to errors in boundary areas. Therefore, we utilize depth information to refine the masks. For each boundary point of the original mask, we examine a five-pixel radius area centered around it, calculating the range of depth values for the pixels within the mask in this region. For the part inside the mask in this region, we calculate the range of depth values of all pixels. For pixels outside the mask in this area, those with depth values within the calculated range are considered part of the mask and subsequently integrated.

\noindent\textbf{Background inpainting.} For the removed dynamic objects, we repair the occluded backgrounds using static information obtained from previous viewpoints, synthesizing a realistic image without dynamic objects. The repaired image contains more scene information, making the appearance of the map more accurate and enhancing the stability of camera tracking. Utilizing the known positions of both previous and current frames, we project a sequence of prior keyframes onto the segmented area's RGB and depth images in the current frame. Some regions remain unfilled because these areas either have not yet appeared in the scene or appeared but lack valid depth information. Figure \ref{fig:Figure1} shows three frames from our self-captured dataset as input and the final reconstructed scene. It is noticeable that dynamic objects are successfully removed, with most of the segmented parts well repaired. 

\subsection{Mask-guided keyframe selection}
For the tracked input frames, we select a set of keyframes, denoted as $K$. Our preference for keyframes leans towards: 1. Frames with a lower ratio of dynamic objects; 2. Frames with a lower overlap ratio with the previous keyframe. We use $I.R_d$ and $I.R_o$ to represent the two ratios respectively for an input frame $I$. When the sum of these two ratios is less than threshold $\tau_2$, we insert the current frame into the keyframe set. To address the inaccuracies and missing information in background inpainting, we reduce the proportion of dynamic objects within keyframes. This approach ensures integration of more reliable information, enhancing the accuracy and stability of camera tracking. Meanwhile, smaller overlaps between keyframes allow keyframe set to include more scene information. In static scenes, this strategy defaults to selection based on overlap ratio.

When selecting keyframes from $K$ for scene representation optimization, we alternate between coverage-based \cite{14} and overlap-based \cite{10} strategies, aiming to strike a balance between optimization efficiency and quality. The coverage-based strategy prefers frames covering the largest scene area, ensuring comprehensive optimization, particularly of edge regions. However, this approach method often requires numerous iterations to optimize relatively small edge areas, diminishing the overall optimization efficiency. It also leads to repetitive selection results, as the coverage area of frames is constant, and frames with large coverage areas maintain higher priority. The overlap-based strategy involves random selection from keyframes that visually overlap with the current frame. To avoid excessive focus on edge regions and recurrent optimization of the identical area, we first optimize the entire scene using the coverage-based strategy, then employ the overlap-based strategy multiple times, periodically repeating this process.

\subsection{Scene representation \& image rendering}
\noindent\textbf{Scene representation.} For scene representation, we adopt multiresolution geometric feature grid $G_\alpha = \{G_\alpha^l\}_{l=1}^L$ and color feature grid $C_{\phi}$, where $l \in \{0, 1, 2\}$ is referred to coarse, mid and fine-level scene details. Feature vectors $G_\alpha\left(\mathbf{x}\right)$ and $C_{\phi}\left(\mathbf{x}\right)$ at each sampled point $\mathbf{x}$ are queried via trilinear interpolation. Each feature grid corresponds to an MLP decoder, with the geometric decoder denoted as $f^{l}$ and the color decoder denoted as $g$. The geometry decoder outputs the predicted occupancy $o_\mathbf{x}$ as:
\begin{equation}
	{o^{l}_\mathbf{x}}=f^{l}\left(\mathbf{x},G_\alpha^l\left(\mathbf{x}\right)\right), {o_\mathbf{x}}=\sum_{l=1}^Lo^{l}_\mathbf{x}.
	\label{eq:1}
\end{equation}
We also encode the color information allowing us to render RGB images which provides additional signals for tracking. The color decoder predicts the color value $c_\mathbf{x}$ as:
\begin{equation}
	{c_\mathbf{x}}=g_\phi\left(\mathbf{x},C_\phi\left(\mathbf{x}\right)\right).
	\label{eq:2}
\end{equation}
Here $\theta = \{\alpha, \phi\}$ are the learnable parameters for geometric and color feature grids.

\noindent\textbf{Image rendering.} Given the camera pose, we can compute the viewing direction $\mathbf{r}$ at pixel coordinates. We sample rays at each pixel and subsequently sample $M$ points $\mathbf{x}_i = \mathbf{o} + d_i\mathbf{r}, i \in \{1, \dots, M\}$ along each ray, where $\mathbf{o}$ represents the camera origin and $d_i$ denotes to the depth value of $\mathbf{x}_i$. Following \cite{9, 10}, for each sampled point, after obtaining the predicted color $c_i$, color and depth can be rendered as
\begin{equation}
	\hat{\mathbf{C}}=\frac{1}{\sum_{i=1}^M w_i} \sum_{i=1}^M w_i \mathbf{c}_i, \quad \hat{D}=\frac{1}{\sum_{i=1}^M w_i} \sum_{i=1}^M w_i d_i,
	\label{eq:3}
\end{equation}

\noindent where $w_i=o_{\mathbf{x}_i} \prod_{j=1}^{i-1}\left(1-o_{\mathbf{x}_j}\right)$ is the computed weight along the ray.

For ray-wise sampling, alongside $M_\text{strat}$ points for stratified sampling, we use surface sampling
, specifically targeting $M_\text{surf}$ points within a distance of less than a threshold $\tau_3$ from surface. In total, we sample $M=M_\text{strat}+M_\text{surf}$ points for each ray. We then verify whether these sampling points fall within valid feature grids. Points outside these grids are excluded, as they contribute no value to the rendering process, enhancing rendering efficiency.

\subsection{Mapping and tracking}
In mapping process, we sample $N$ pixels from the selected keyframes to optimize scene representation. Subsequently, we adopt a staged approach to optimization, aiming to minimize both geometric and photometric losses. 

Geometric loss and photometric loss are applied as the $L_1$ loss between the predicted and ground truth values for color and depth, respectively, as follows:
\begin{equation}
	\mathcal{L}_g=\frac{1}{N} \sum_{n=1}^N\left|D_n-\hat{D}_n\right|, \mathcal{L}_p=\frac{1}{N} \sum_{n=1}^N\left|C_n-\hat{C}_n\right|.
	\label{eq:4}
\end{equation}

We jointly optimize feature $\theta$ and the camera extrinsic parameters $\{\mathbf{R}_i, \mathbf{t}_i\}$ of $K$ selected keyframes as
\begin{equation}
	\min _{\theta,\left\{\mathbf{R}_i, \mathbf{t}_i\right\}}\left(\mathcal{L}_g+\lambda_p \mathcal{L}_p\right), 
	\label{eq:5}
\end{equation}
where $\lambda_p$ is the loss weighting factor. 

In parallel, we run a tracking process, sampling $N_t$ pixels from the current frame and applying the same $\mathcal{L}_{g\_var}$ as in \cite{10} to optimize the camera poses $\{\mathbf{R}, \mathbf{t}\}$ of the current frame as:
\begin{equation}
	\min _{\mathbf{R}, \mathbf{t}}\left(\mathcal{L}_{g\_var}+\lambda_{pt} \mathcal{L}_p\right) .
	\label{eq:6}
\end{equation}
Overall, our algorithmic framework is shown in Algorithm~\ref{alg:Algorithm1}.

\begin{algorithm}[!ht]
	\small\caption{Framework of NID-SLAM}
	\label{alg:Algorithm1}
	\renewcommand{\algorithmicrequire}{\textbf{Input:}}
	\renewcommand{\algorithmicensure}{\textbf{Output:}}
	\begin{algorithmic}[1]
		\small
		\REQUIRE Raw RGB-D frames: \{$I_1$, \dots, $I_n$\}
		\ENSURE Rendered RGB-D frames :\{$R_1$, \dots, $R_n$\}
		
		\FOR {each frame $I_i$}
		\STATE Remove dynamic objects in $I_i$
		
		\IF {$I_i.R_d + I_i.R_o < \tau_2$}
		\STATE Insert $I_i$ into keyframe set $K$
		\ENDIF
		
		\STATE Select frames for optimization from $K$
		\STATE Optimize $G_\alpha$ and $C_{\phi}$ using selected frames
		\STATE Predict color $c_\mathbf{x}$ for sampled points $\mathbf{x}$ by \eqref{eq:2}
		\STATE Render $R_i$ by \eqref{eq:3}
		\ENDFOR
	\end{algorithmic}
\end{algorithm}

\section{Experiments}
\begin{table*}[!t]
	\begin{center}
		\small
		\caption{Camera tracking results on TUM RGB-D. ATE RMSE $[m](\downarrow)$ is used as evaluation metric. \textbackslash{} represents corresponding data is not mentioned in the respective literature.}\label{tab:Table 1.}
		\resizebox{\linewidth}{!}{
			\begin{tabular}{lcccccccc}
				\toprule
				\textbf{} & \text{fr3/w/half} & \text{fr3/w/xyz} & \text{fr3/w/rpy} & \text{fr3/w/static} 
				& \text{fr3/s/half} & \text{fr3/s/xyz} & \text{fr3/s/rpy} & \text{fr3/s/static}\\
				\midrule
				iMAP$^\star$ \cite{9} & 0.638 & 0.410 & 0.834 & 0.102
				& 0.812 & 0.420 & 0.364 & 0.074\\
				NICE-SLAM \cite{10} & 0.629 & 0.302 & 0.724 & 0.092 
				& 0.569 & 0.394 & 0.099 & 0.029\\
				\textbf{NID-SLAM(Ours)} & 0.071 & 0.064 & 0.648 & 0.062
				& 0.105 & 0.075 & 0.086 & 0.019\\
				\midrule
				ORB-SLAM2 \cite{1} & 0.467 & 0.721 & 0.784 & 0.387 
				& \textbf{0.019} & \textbf{0.009} & \textbf{0.019} & \textbf{0.009}\\
				Dyna-SLAM \cite{3} & \textbf{0.029} & \textbf{0.015} & \textbf{0.136} & \textbf{0.007} 
				& 0.025 & 0.013 & \textbackslash{} & \textbackslash{}\\
				\bottomrule
		\end{tabular}}
	\end{center}
\end{table*}

\begin{table}[!t]
	\begin{center}
		\small
		\caption{\textit{Translational} RPE RMSE results $[m](\downarrow)$ on TUM.}\label{tab:Table 3.}
		\resizebox{\linewidth}{!}{
			\begin{tabular}{lcccc}
				\toprule
				\textbf{} & \text{fr3/w/half} & \text{fr3/w/xyz} & \text{fr3/s/half} & \text{fr3/s/xyz} 
				\\
				\midrule
				iMAP$^\star$ \cite{9} & 0.990 & 0.656 & 1.278 & 0.787
				\\
				NICE-SLAM \cite{10} & 0.978 & 0.493 & 0.925 & 0.890
				\\
				\textbf{NID-SLAM} & 0.105 & 0.096 & 0.149 & 0.114
				\\
				\midrule
				ORB-SLAM2 \cite{1} & 0.348 & 0.394 & \textbf{0.023} & \textbf{0.012} 
				\\
				Dyna-SLAM \cite{3} & \textbf{0.028} & \textbf{0.022} & 0.024 & 0.014 
				\\
				\bottomrule
		\end{tabular}}
	\end{center}
\end{table}

\begin{table}[t]
	\begin{center}
		\small
		\caption{\textit{Rotational} RPE RMSE results $[m](\downarrow)$ on TUM.}\label{tab:Table 5.}
		\resizebox{\linewidth}{!}{
			\begin{tabular}{lcccc}
				\toprule
				\textbf{} & \text{fr3/w/half} & \text{fr3/w/xyz} & \text{fr3/s/half} & \text{fr3/s/xyz} 
				\\
				\midrule
				iMAP$^\star$ \cite{9} & 96.650 & 36.240 & 75.195 & 25.547
				\\
				NICE-SLAM \cite{10} & 81.731 & 15.286 & 73.552 & 39.729
				\\
				\textbf{NID-SLAM} & 3.227 & 2.114 & 4.922 & 2.760
				\\
				\midrule
				ORB-SLAM2 \cite{1} & 7.214 & 7.785 & \textbf{0.601} & \textbf{0.489} \\
				Dyna-SLAM \cite{3} & \textbf{0.784} & \textbf{0.628} & 0.705 & 0.504\\
				\bottomrule
		\end{tabular}}
	\end{center}
\end{table}

\begin{figure*}[t!]
	\centering
	\begin{minipage}[c]{0.03\linewidth}
		\centering
		\rotatebox{90}{\small\texttt{fr3/w/xyz}}
	\end{minipage}%
	\begin{minipage}[c]{0.18\linewidth}
		\centering\small\texttt{Sequence}
		{\epsfig{figure=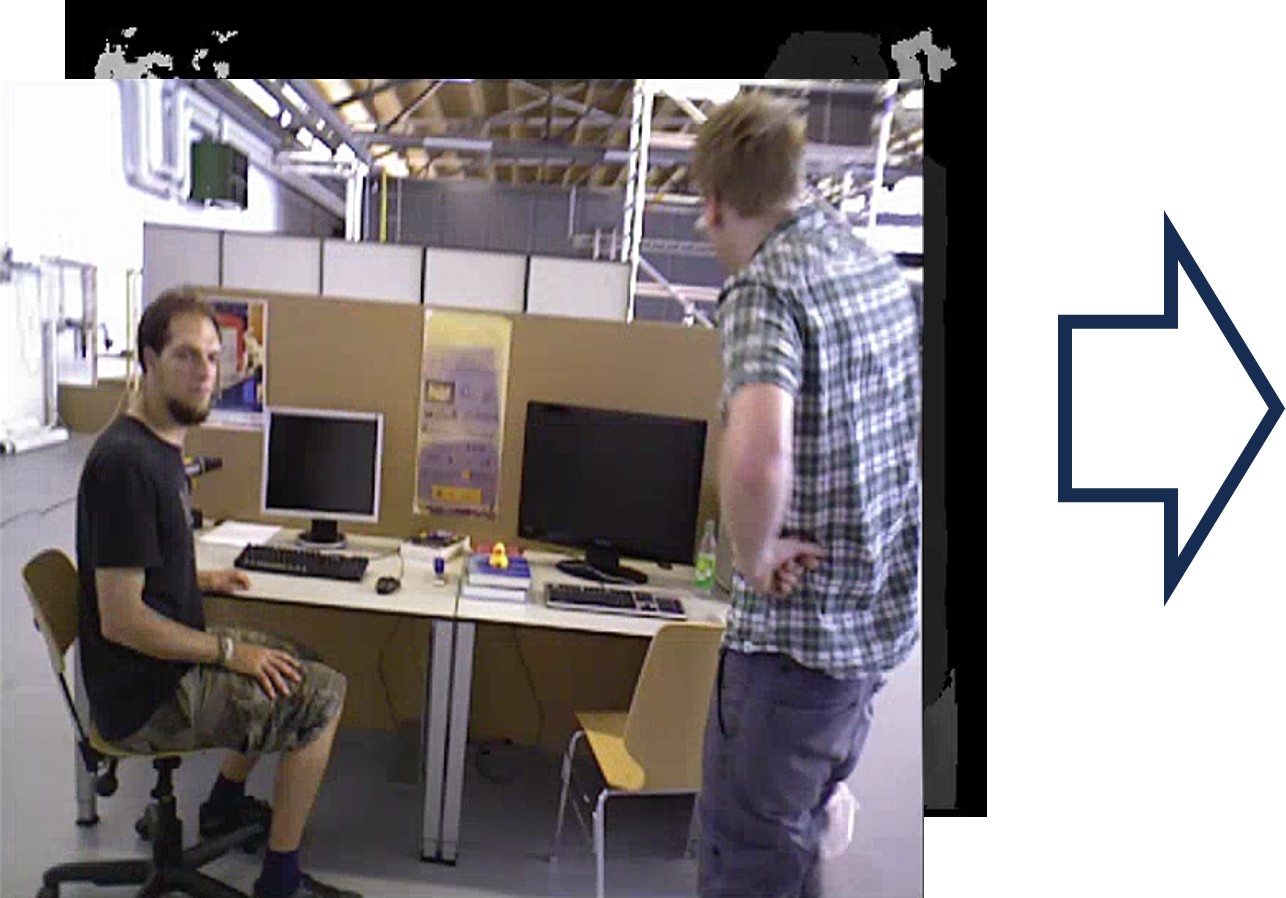,width=3cm}}
	\end{minipage}
	\begin{minipage}[c]{0.18\linewidth}
		\centering\small\texttt{NICE-SLAM~w/o~mask}
		{\epsfig{figure=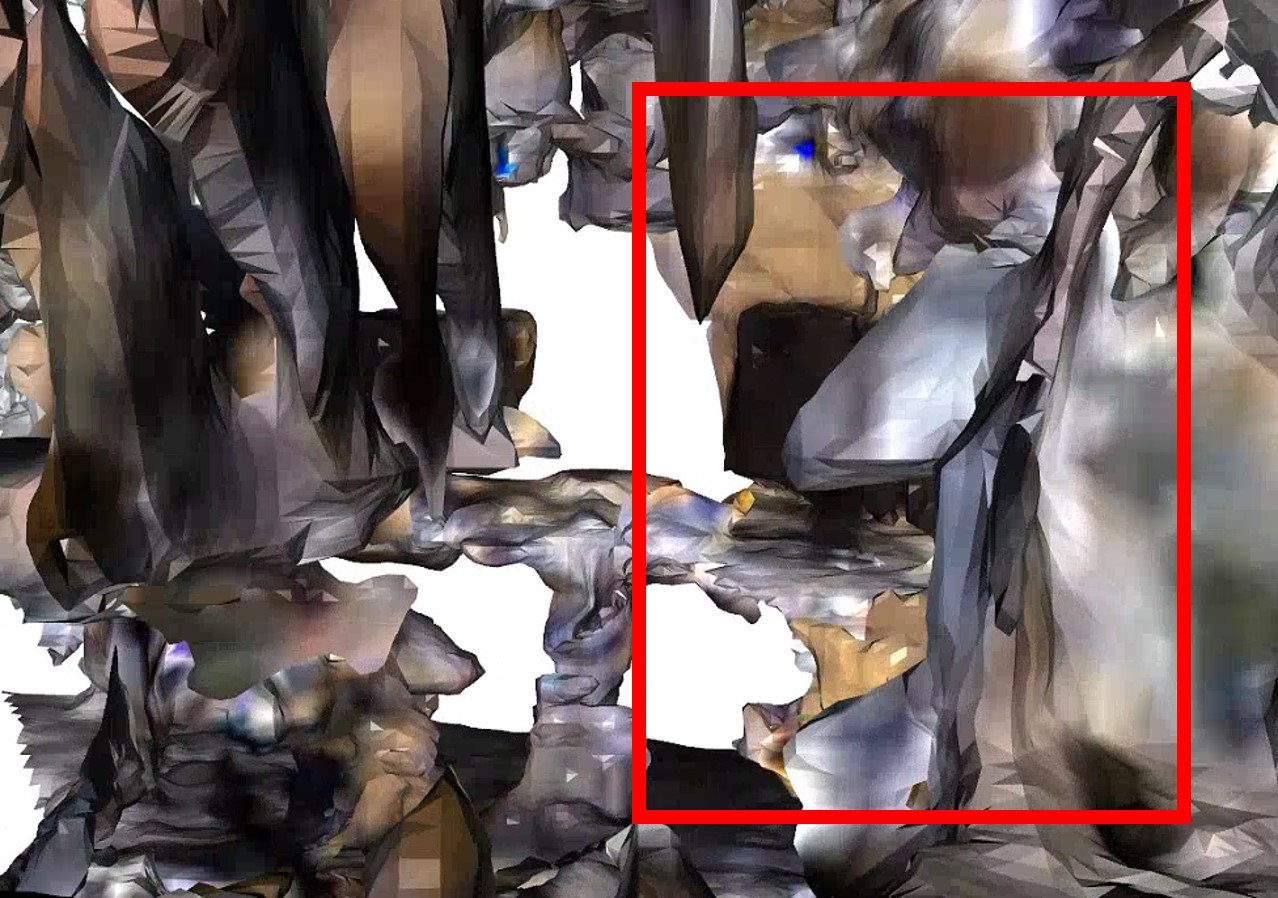,width=3cm}}
	\end{minipage}
	\vspace{2pt}
	\begin{minipage}[c]{0.18\linewidth}
		\centering\small\texttt{iMAP$^\star$~w~mask}
		{\epsfig{figure=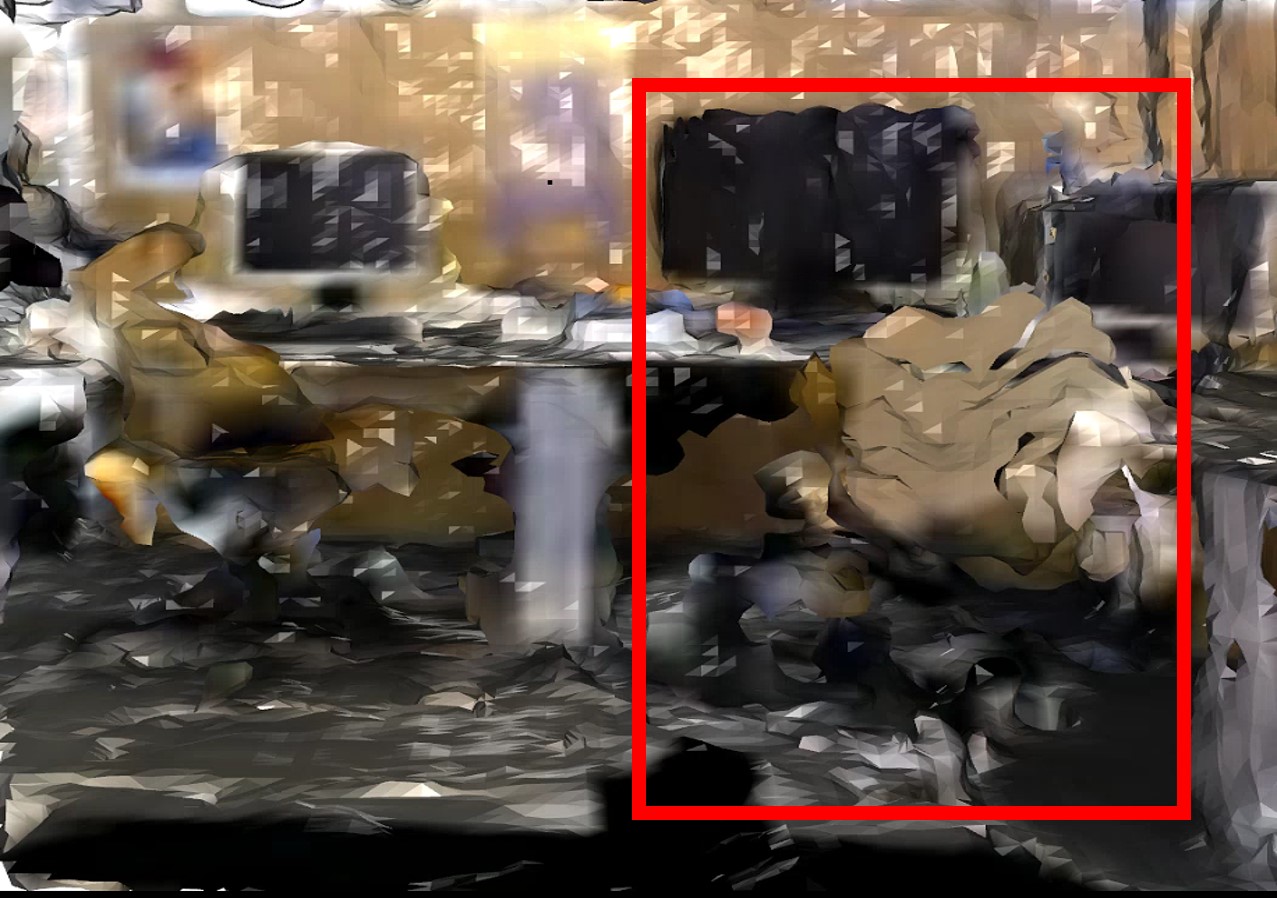,width=3cm}}
	\end{minipage}
	\vspace{2pt}
	\begin{minipage}[c]{0.18\linewidth}
		\centering\small\texttt{NICE-SLAM~w~mask}
		{\epsfig{figure=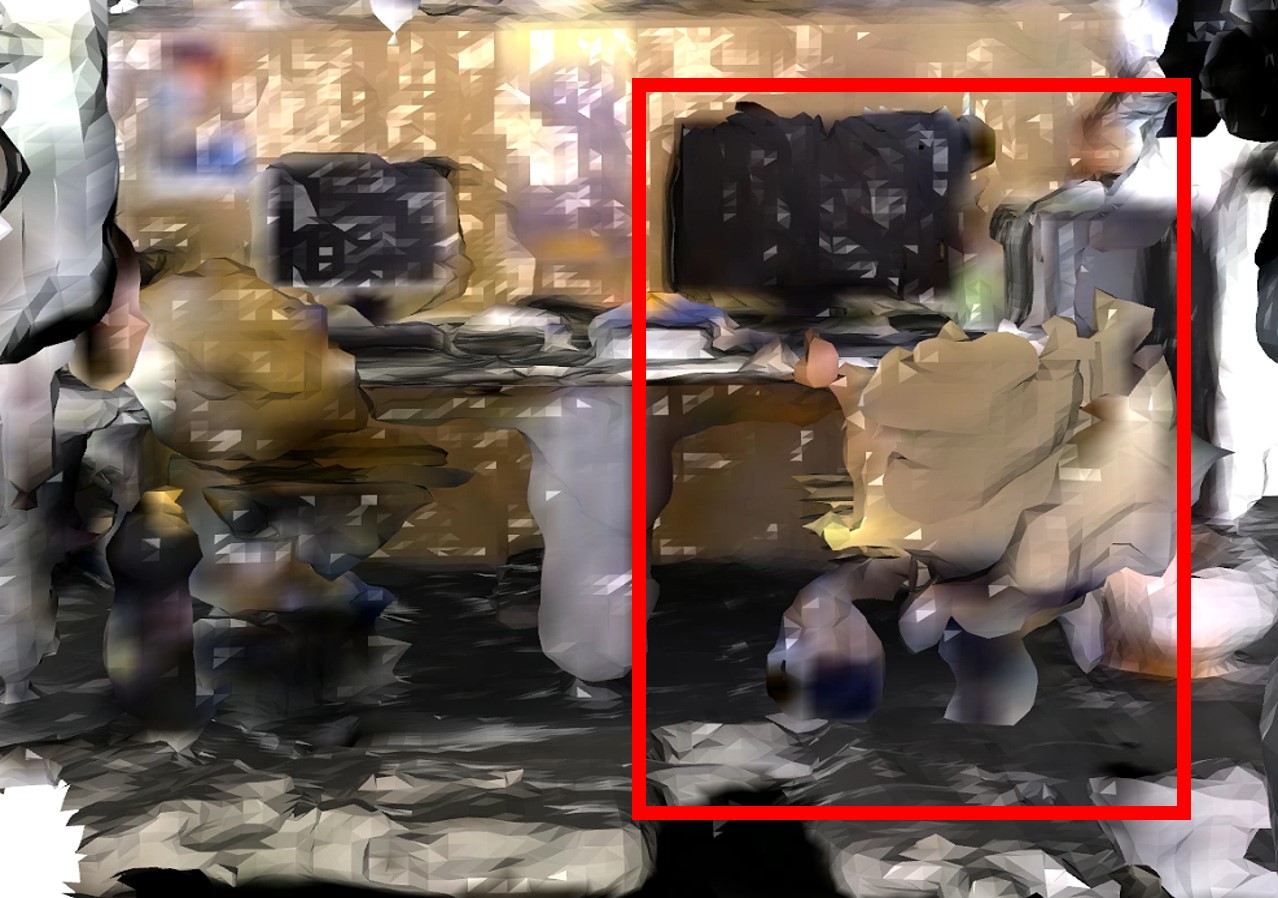,width=3cm}}
	\end{minipage}
	\vspace{2pt}
	\begin{minipage}[c]{0.18\linewidth}
		\centering\small\texttt{NID-SLAM(Ours)}
		{\epsfig{figure=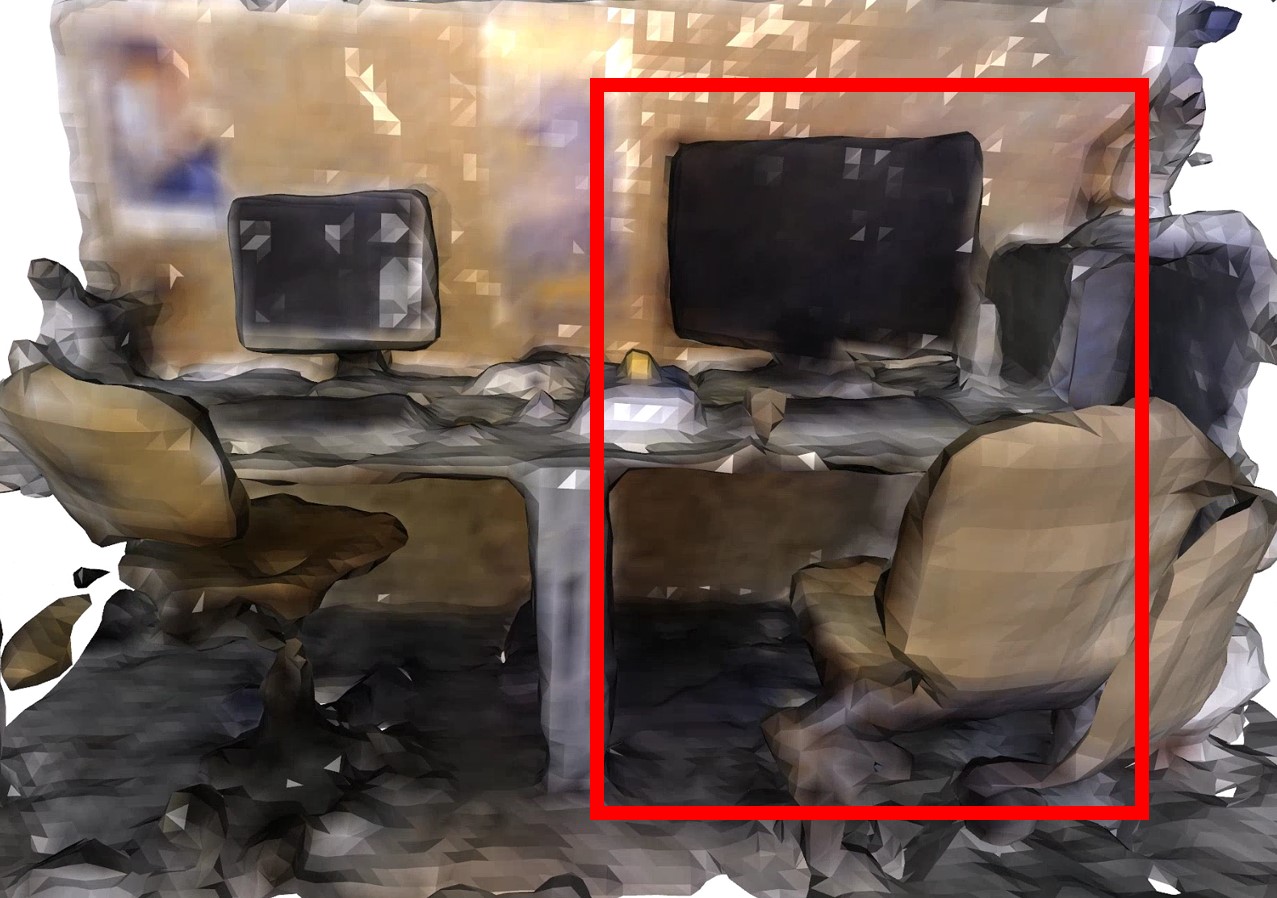,width=3cm}}
	\end{minipage}
	\vspace{2pt}
	\centering
	\begin{minipage}[c]{0.03\linewidth}
		\centering
		\rotatebox{90}{\small\texttt{fr3/s/half}}
	\end{minipage}%
	\begin{minipage}[c]{0.18\linewidth}
		\centering
		{\epsfig{figure=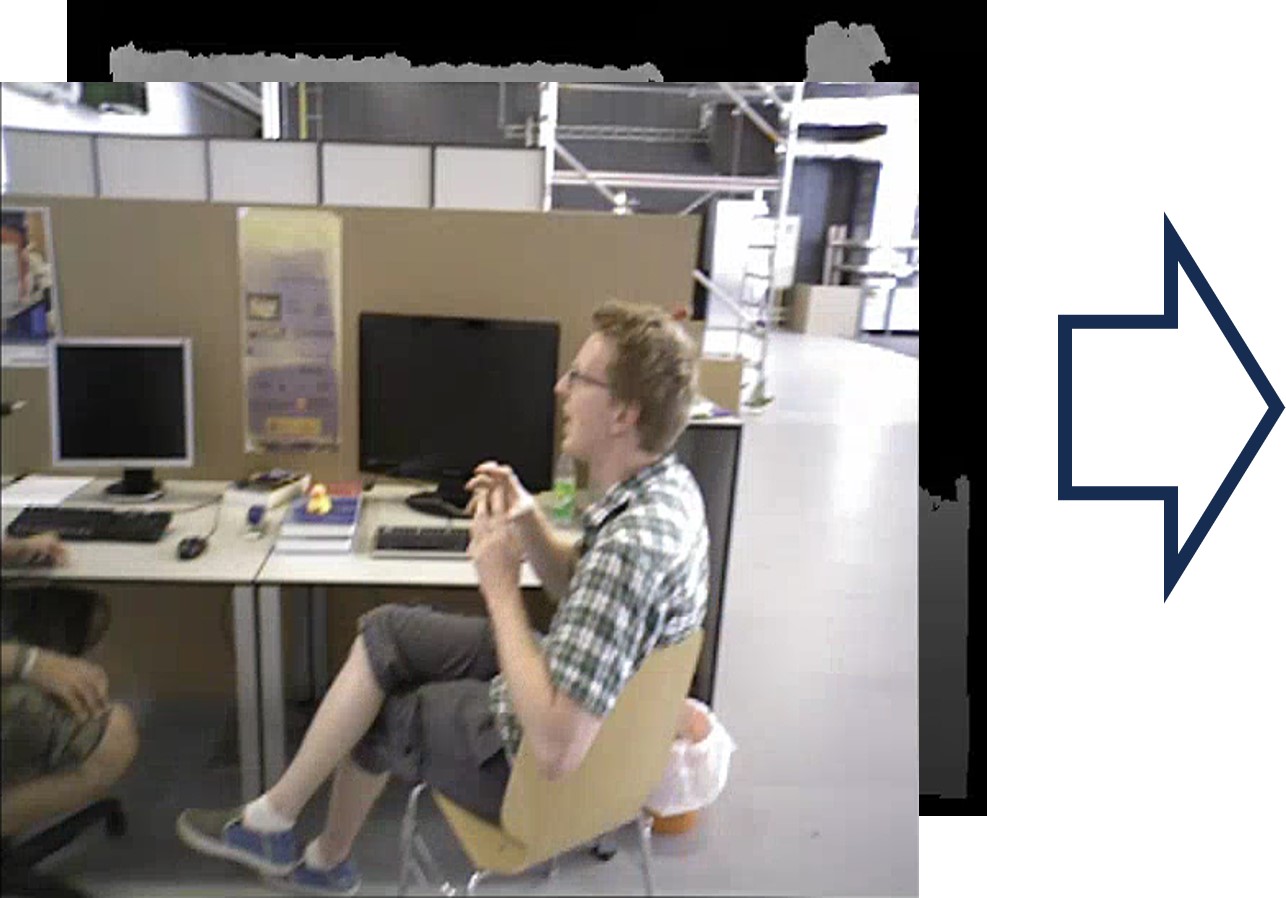,width=3cm}}
	\end{minipage}
	\begin{minipage}[c]{0.18\linewidth}
		\centering
		{\epsfig{figure=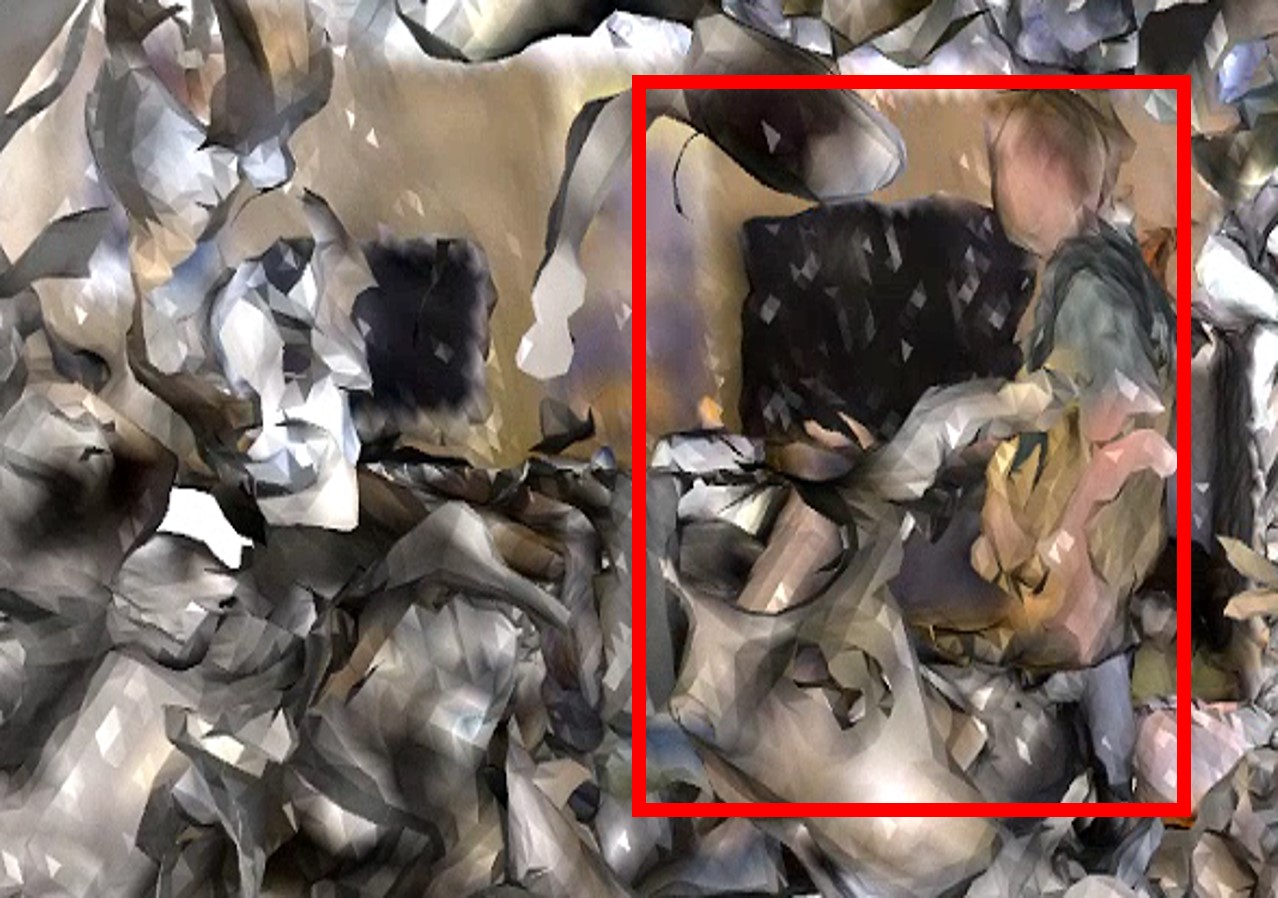,width=3cm}}
	\end{minipage}
	\vspace{2pt}
	\begin{minipage}[c]{0.18\linewidth}
		\centering
		{\epsfig{figure=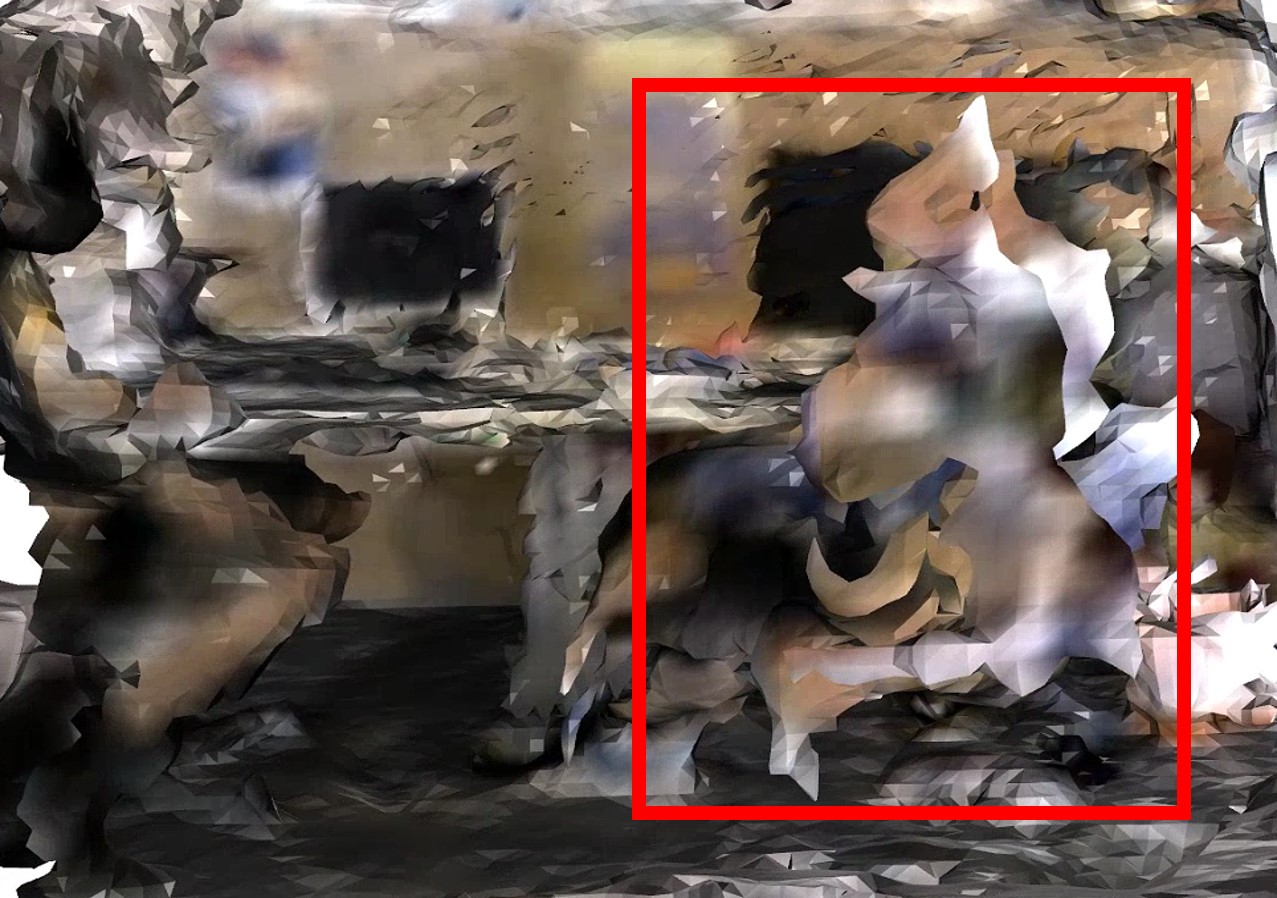,width=3cm}}
	\end{minipage}
	\vspace{2pt}
	\begin{minipage}[c]{0.18\linewidth}
		\centering
		{\epsfig{figure=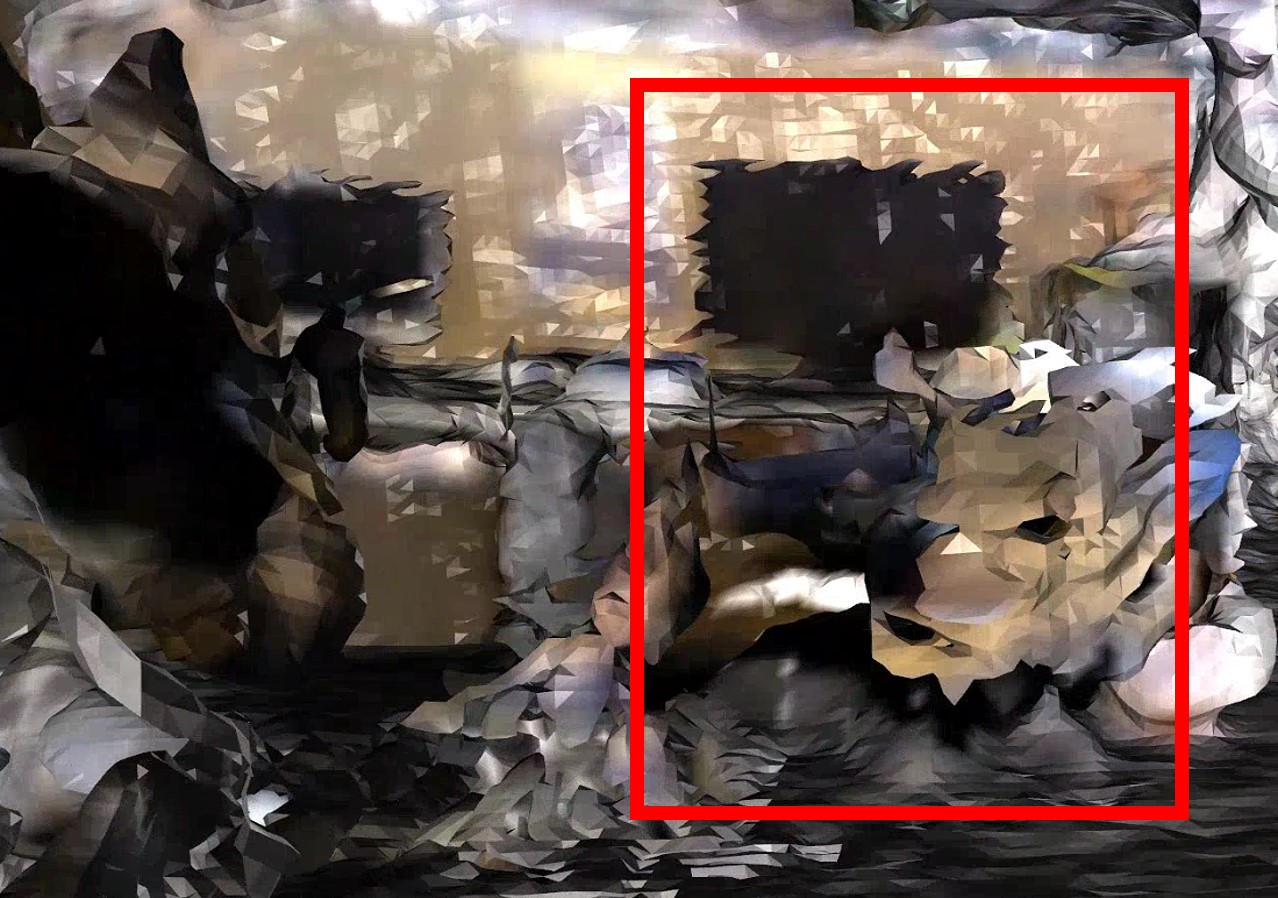,width=3cm}}
	\end{minipage}
	\vspace{2pt}
	\begin{minipage}[c]{0.18\linewidth}
		\centering
		{\epsfig{figure=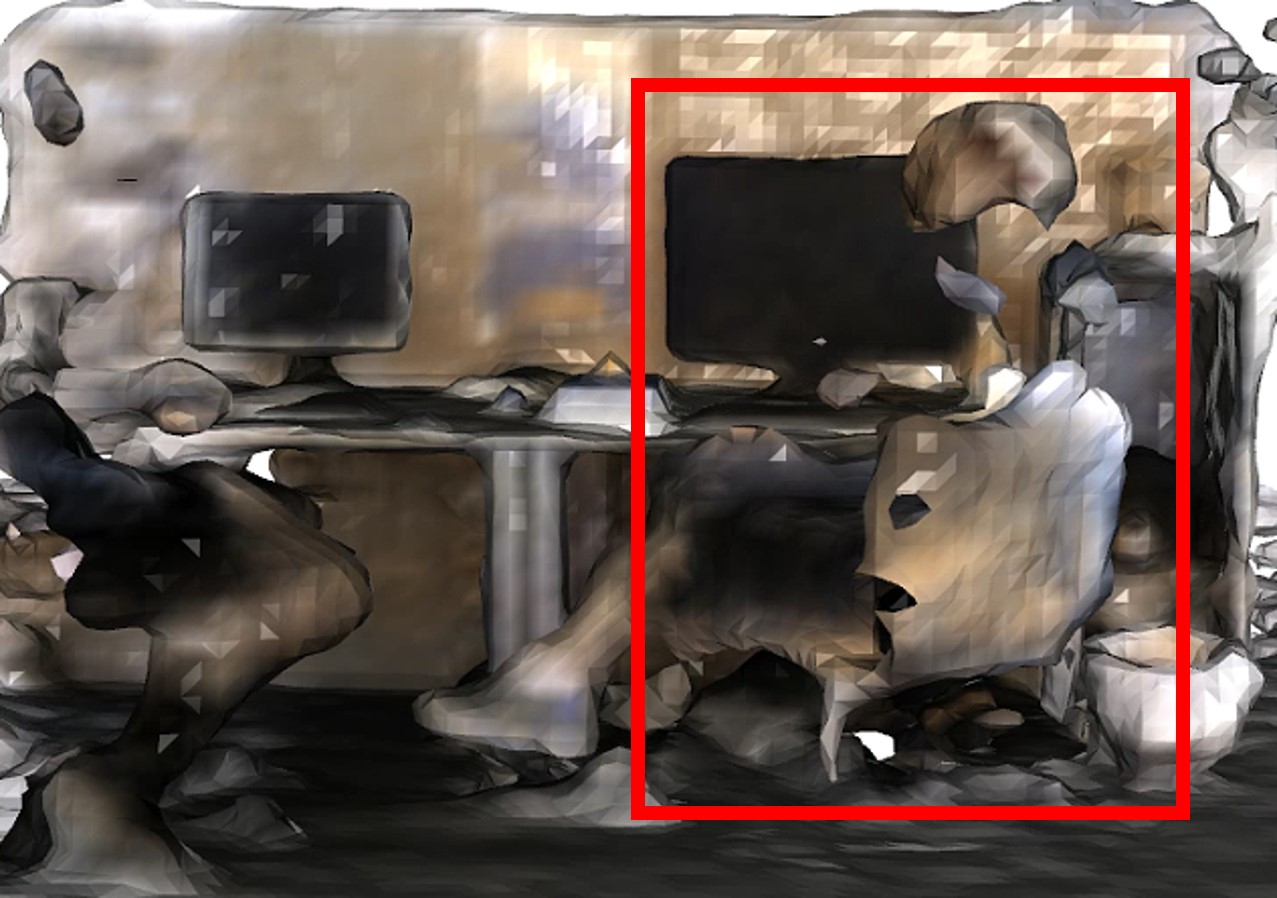,width=3cm}}
	\end{minipage}
	\vspace{2pt}
	\centering
	\begin{minipage}[c]{0.03\linewidth}
		\centering
		\rotatebox{90}{\small\texttt{fr3/s/static}}
	\end{minipage}%
	\begin{minipage}[c]{0.18\linewidth}
		\centering
		{\epsfig{figure=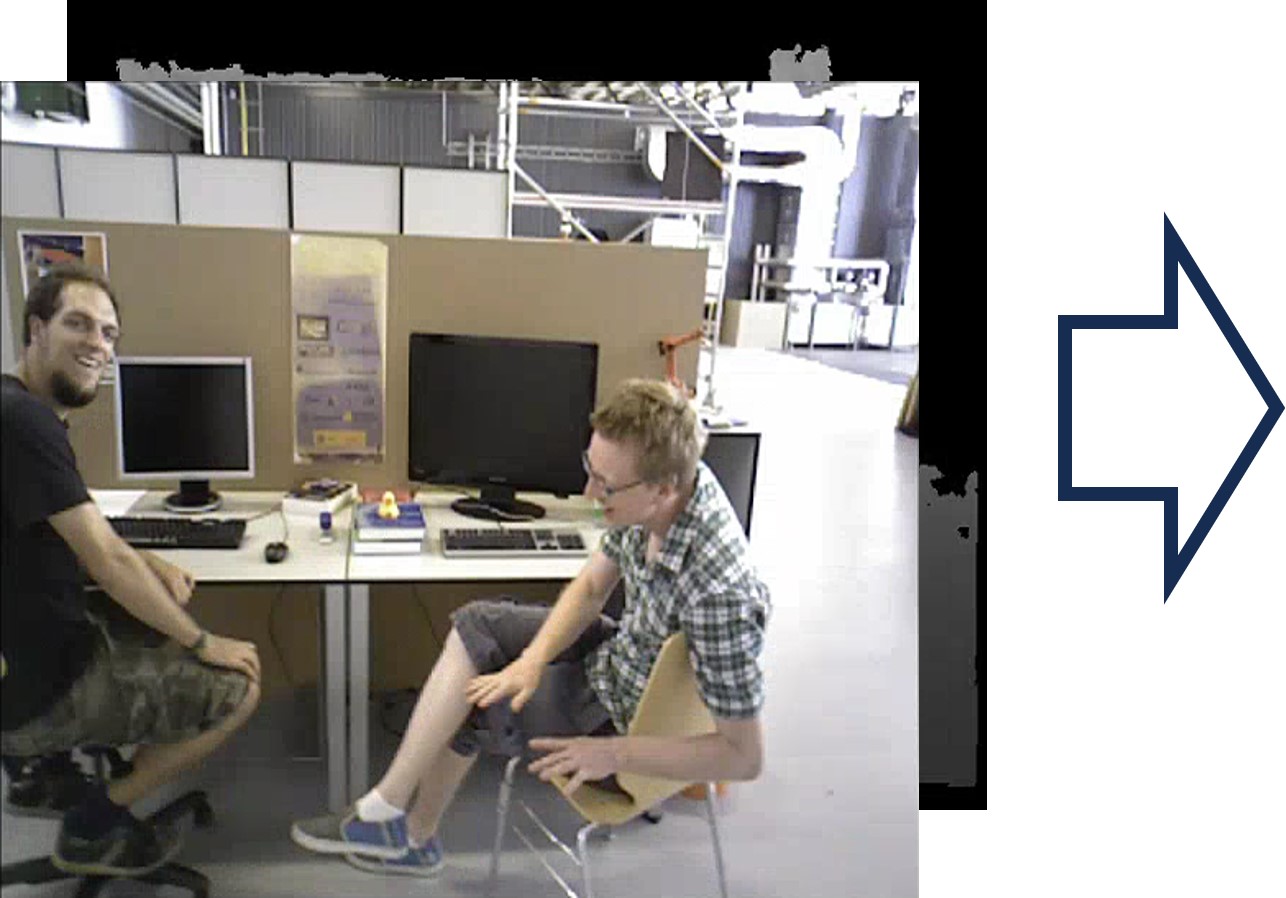,width=3cm}}
	\end{minipage}
	\begin{minipage}[c]{0.18\linewidth}
		\centering
		{\epsfig{figure=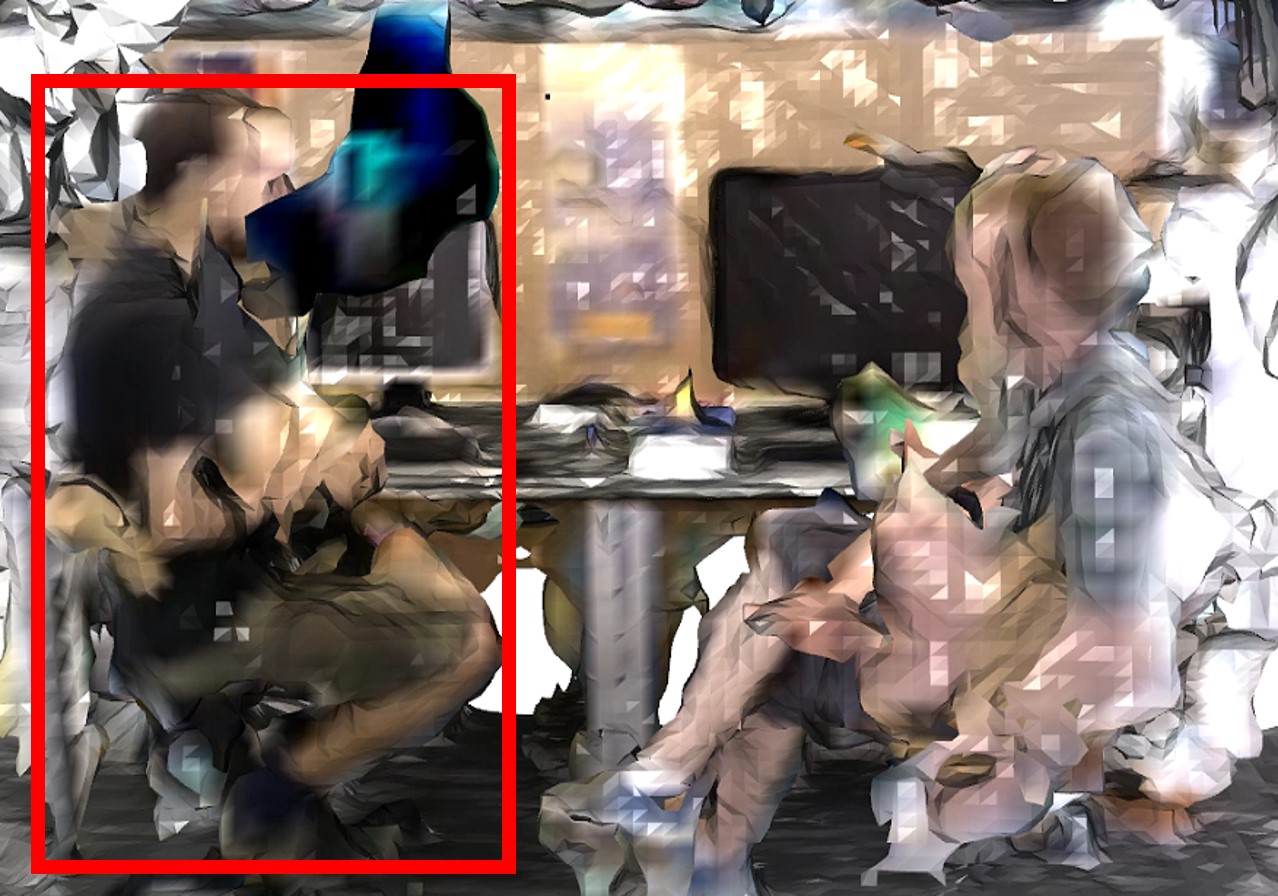,width=3cm}}
	\end{minipage}
	\begin{minipage}[c]{0.18\linewidth}
		\centering
		{\epsfig{figure=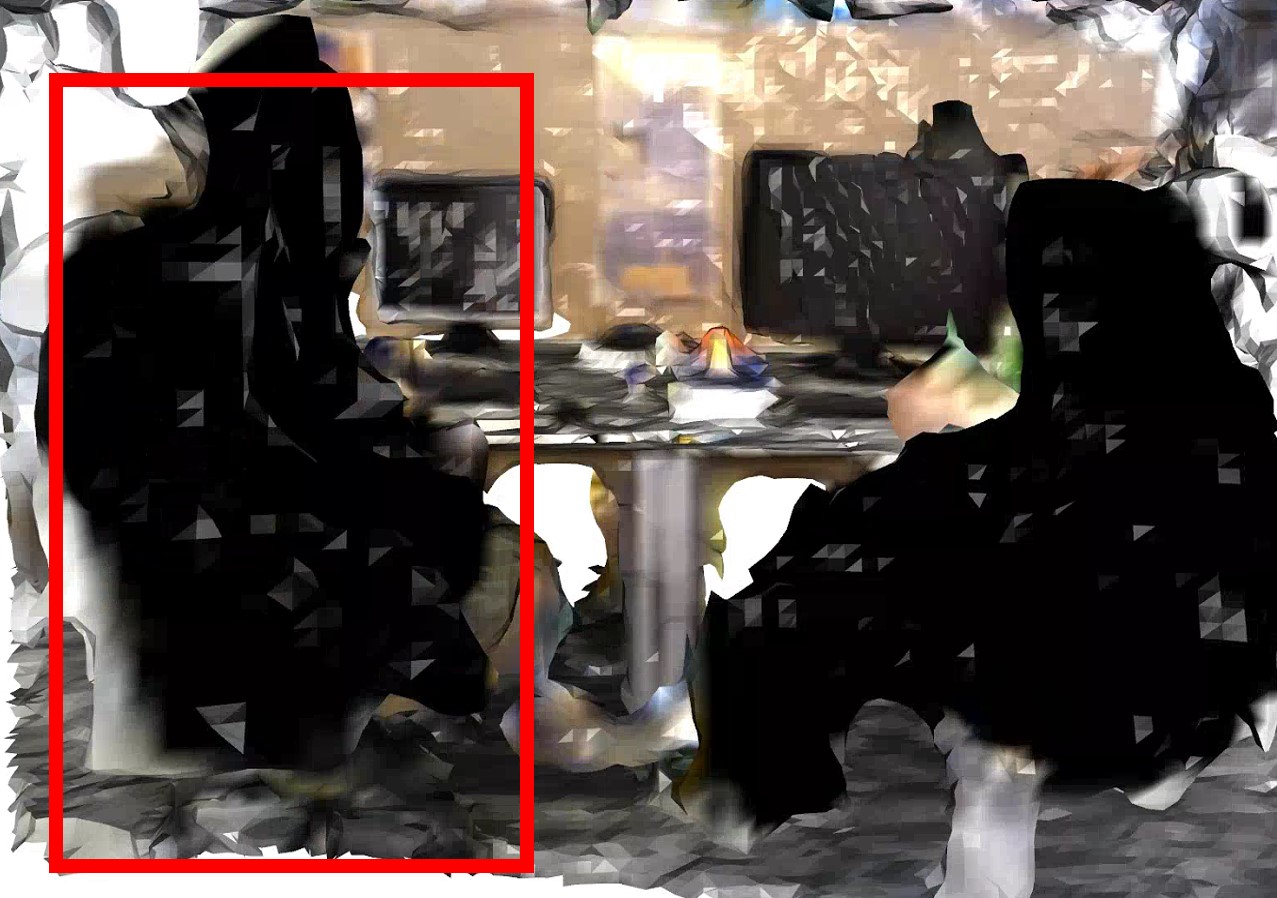,width=3cm}}
	\end{minipage}
	\begin{minipage}[c]{0.18\linewidth}
		\centering
		{\epsfig{figure=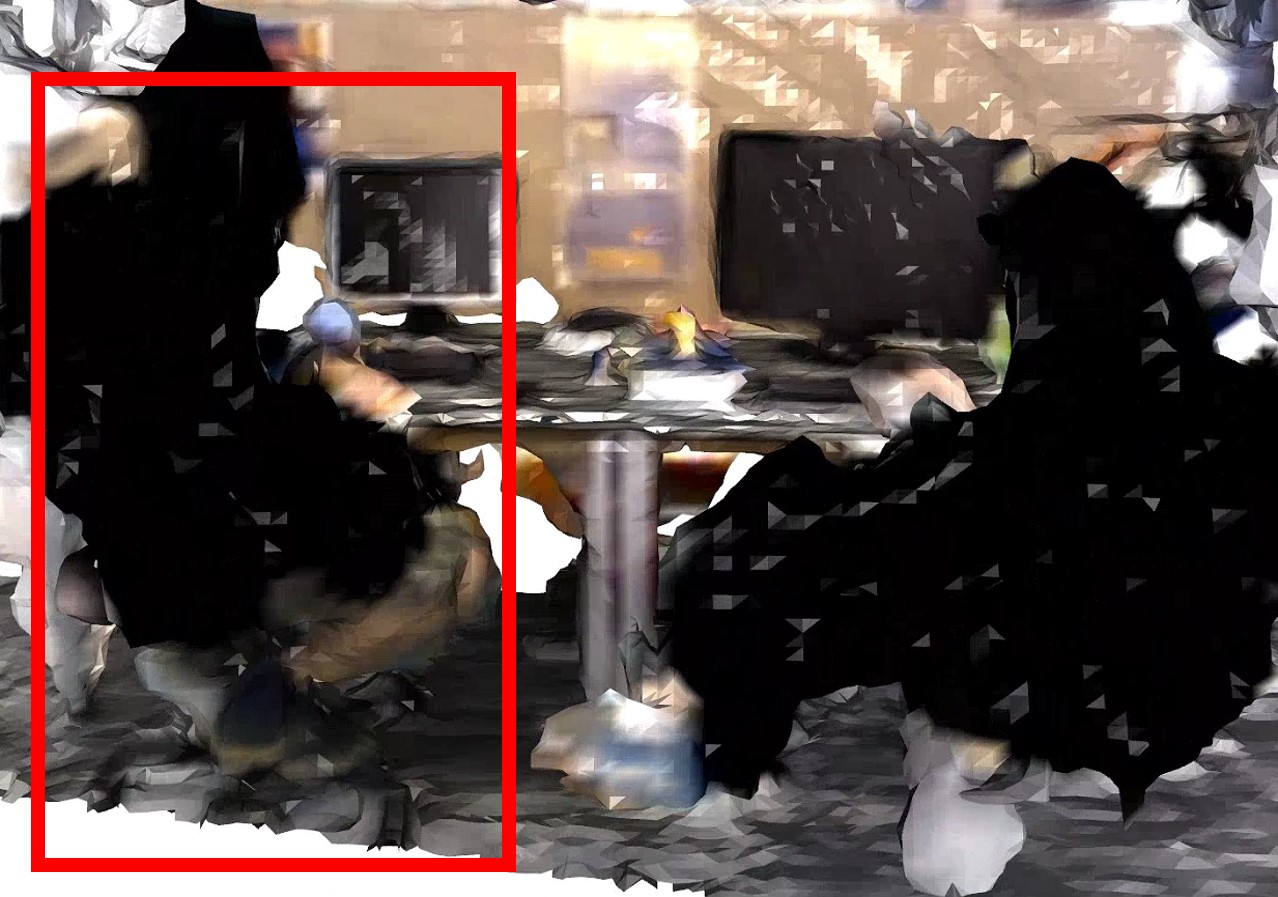,width=3cm}}
	\end{minipage}
	\begin{minipage}[c]{0.18\linewidth}
		\centering
		{\epsfig{figure=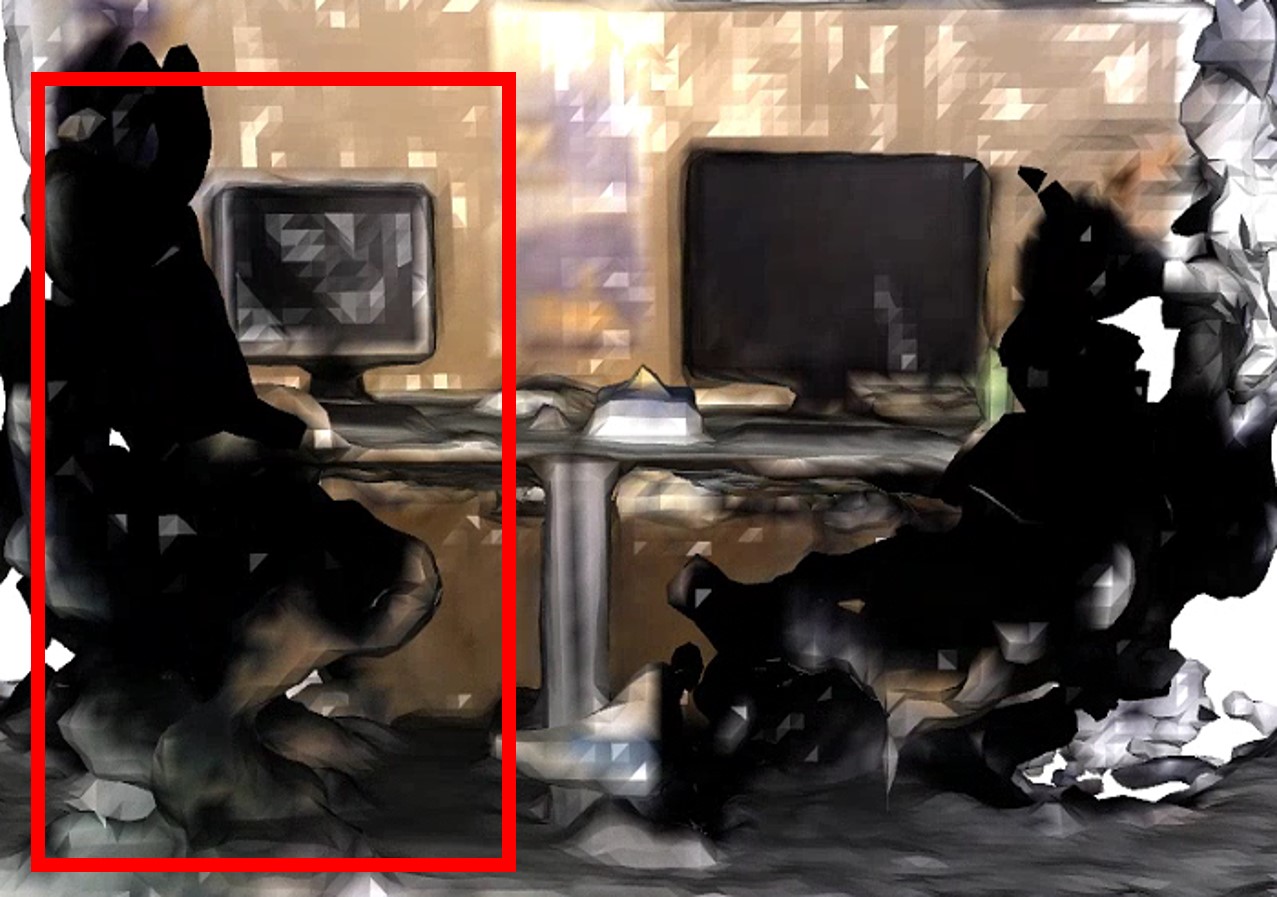,width=3cm}}
	\end{minipage}
	\caption{Reconstruction results on TUM RGB-D. The red box highlights areas with dynamic object.}
	\label{fig:Figure3}
\end{figure*} 

\subsection{Experimental setup}
\noindent\textbf{Datasets.} We evaluate NID-SLAM on a variety of scenes from three different datasets. For dynamic environments, we evaluate pose estimation results on 8 scenes from TUM RGB-D dataset. We follow the same pre-processing step for TUM RGB-D as in \cite{15}. We also evaluate on a self-cpatured room as a larger dynamic scene. Following iMAP and NICE-SLAM, we quantitatively evaluate the reconstruction quality on 8 synthetic scenes from Replica \cite{16}.

\noindent\textbf{Metrics.} We evaluate camera tracking using Absolute Trajectory Error (ATE) and Relative Pose Error (RPE). For evaluation of the reconstruction quality, we adopt \textit{Depth L1} (cm), \textit{Accuracy} (cm), \textit{Completion} (cm), and \textit{Completion ratio} (\%) with a threshold of 5cm. 


\noindent\textbf{Implementation details.} We run NID-SLAM on a desktop PC with a 3.20GHz Intel Core i9-12900KF CPU and NVIDIA RTX 3090 GPU. In all our experiments, we set the number of sampling points on a ray to $M_\text{strat} = 32$ and $M_\text{surf} = 16$; we select $K = 5$ keyframes and sample $N = 1000$ pixels for optimization, along with $N_t = 200$ pixels for tracking. Note that iMAP$^\star$ denotes the iMAP reimplementation released by the NICE-SLAM \cite{10} authors.

\subsection{Evaluation of mapping and tracking}
\noindent{\bf Evaluation on TUM RGB-D \cite{15}.} To evaluate camera tracking performance, we use the dynamic TUM RGB-D dataset. For fair comparison, we provided iMAP$^\star$ and NICE-SLAM with the same dynamic-object-free input as our method. As shown in Table \ref{tab:Table 1.}, our approach achieves the best ATE results among neural SLAM methods and surpasses ORB-SLAM2 in several scenarios. Table \ref{tab:Table 3.} and Table \ref{tab:Table 5.} present the result of translational RPE and rotational RPE, our method significantly outperforms other neural SLAM approaches. However, due to the inherent gap in tracking performance between neural SLAM methods and classic SLAM approaches, NID-SLAM is not yet able to outperform Dyna-SLAM. Nevertheless, our method significantly narrows the performance gap in tracking  between these two categories of methods in dynamic environments. Figure \ref{fig:Figure3} qualitatively presents the reconstruction results on TUM RGB-D. Notably, our approach achieves accurate removal of dynamic objects, significantly improving the reconstruction quality and completeness of dynamic scenes.

\noindent{\bf Evaluation on Replica \cite{16}.} We also evaluate the reconstruction quality on Replica while using the same rendered RGB-D sequence provided by the authors of NICE-SLAM. With the keyframe selection method, our method ensures high-quality reconstruction across both central and edge regions. In Table \ref{tab:Table 2.}, we present a quantitative comparison of the reconstruction and rendering performance of our method and the baselines. The results demonstrate that our approach outperforms the baselines for both 2D and 3D metrics. Qualitatively, we can see from Fig.~\ref{fig:Figure3} that the mesh obtained by our method features notably smoother floors and more details on small objects, such as table lamps and ornaments.

\noindent{\bf Evaluation on a larger dynamic scene.} To evaluate the scalability of our method, we captured a sequence in a large conference room with two people sitting or walking in it. Fig.~\ref{fig:Figure1} shows the reconstruction results obtained using NID-SLAM, demonstrating that our approach's ability to accurately capture geometric details while eliminating holes.

\begin{table}[!t]
	\begin{center}
		\small
		\caption{Reconstruction results in the Replica dataset (average over 8 scenes).}\label{tab:Table 2.}
		\resizebox{\linewidth}{!}{
			\begin{tabular}{lcccc}
				\toprule
				\textbf{} & iMAP$^\star$ & DI-Fusion \cite{17} & NICE-SLAM & Ours\\
				\midrule
				Depth L1 $\downarrow$ & 7.64 & 23.33 & 3.53 & \textbf{2.87}\\
				Acc. $\downarrow$ & 6.95 & 19.40 & 2.85 & \textbf{2.72}\\
				Comp. $\downarrow$ & 5.33 & 10.19 & 3.00 & \textbf{2.56}\\
				Comp. Ratio $\uparrow$ & 66.60 & 72.96 & 89.33 & \textbf{91.16}\\
				\bottomrule
		\end{tabular}}
	\end{center}
\end{table}

\begin{figure}[t!]
	\centering
	\begin{minipage}[c]{0.01\linewidth}
		\centering
		\rotatebox{90}{\tiny\texttt{NICE-SLAM\cite{10}}}
	\end{minipage}%
	\begin{minipage}[c]{.32\linewidth}
		\centering\tiny\texttt{room-0}
		{\epsfig{figure=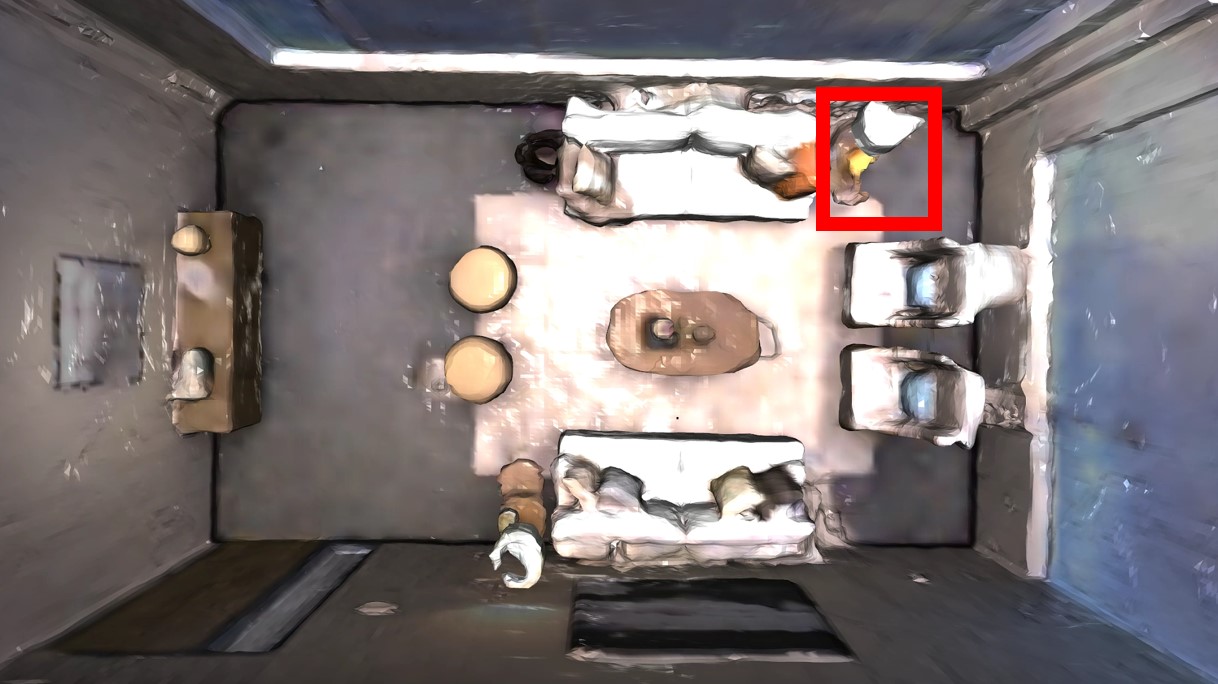,width=2.6cm}}
	\end{minipage}
	\vspace{2pt}
	\begin{minipage}[c]{0.32\linewidth}
		\centering\tiny\texttt{room-1}
		{\epsfig{figure=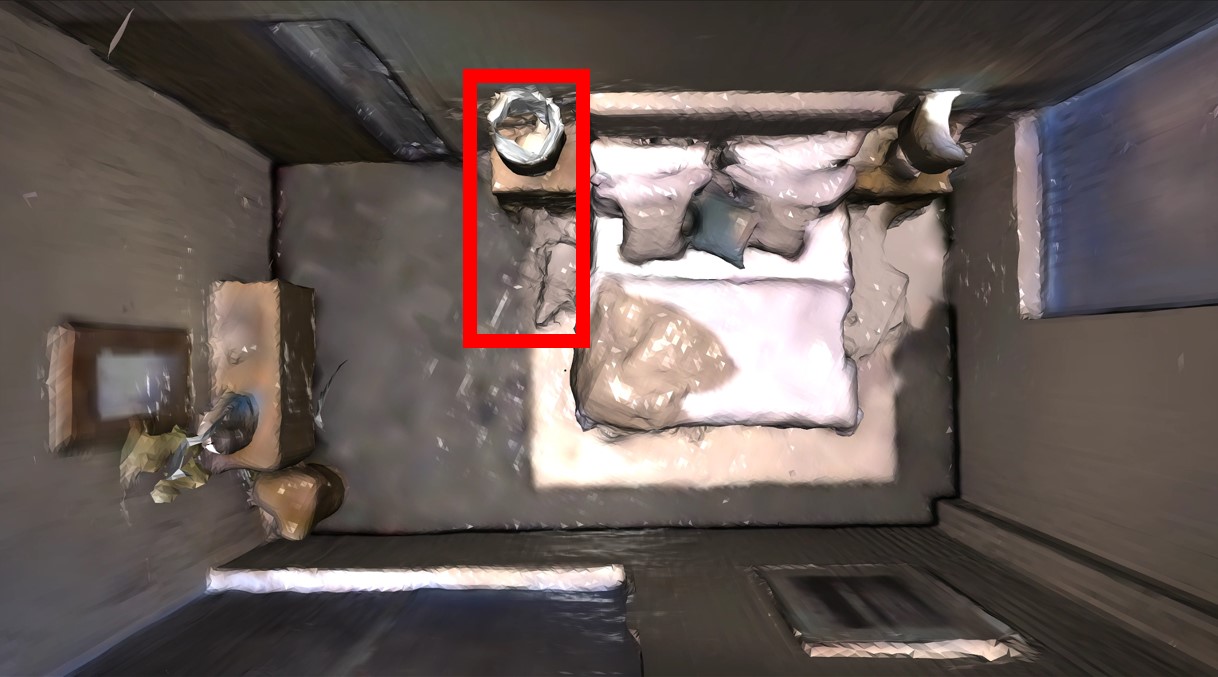,width=2.6cm}}
	\end{minipage}
	\vspace{2pt}
	\begin{minipage}[c]{0.32\linewidth}
		\centering\tiny\texttt{office-2}
		{\epsfig{figure=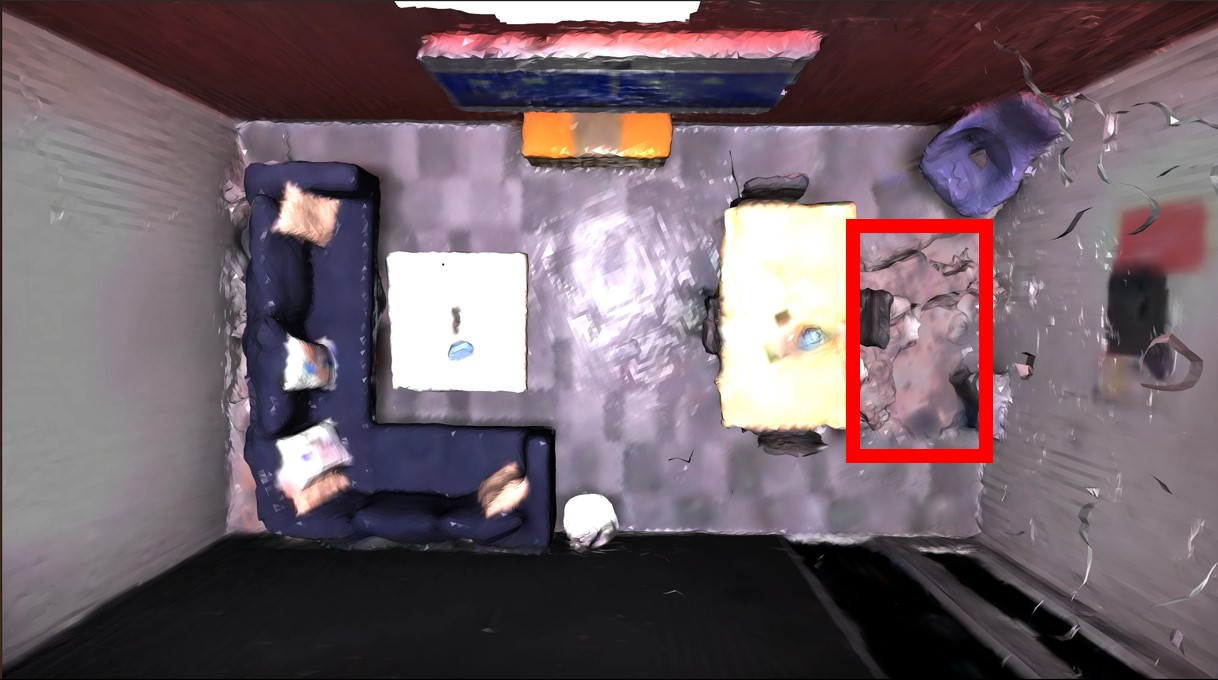,width=2.6cm}}
	\end{minipage}
	\vspace{2pt}
	\centering
	\begin{minipage}[c]{0.01\linewidth}
		\centering
		\rotatebox{90}{\tiny\texttt{Ours}}
	\end{minipage}%
	\begin{minipage}[c]{.32\linewidth}
		\centering
		{\epsfig{figure=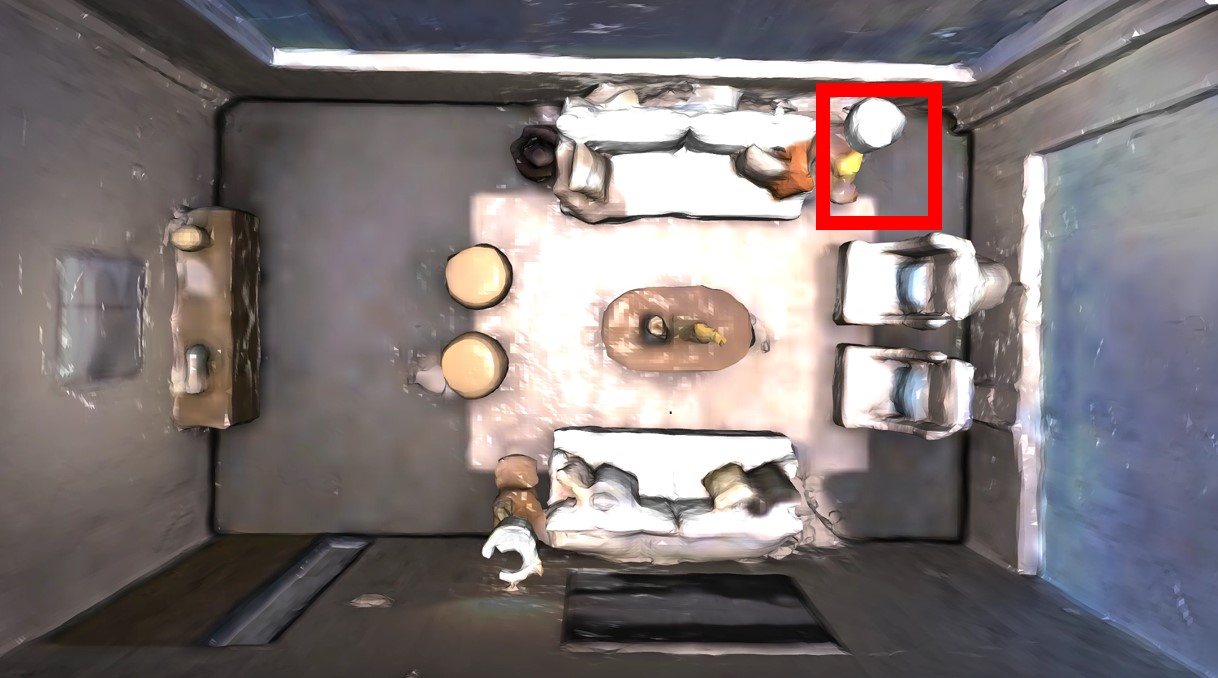,width=2.6cm}}
	\end{minipage}
	\vspace{1pt}
	\begin{minipage}[c]{0.32\linewidth}
		\centering
		{\epsfig{figure=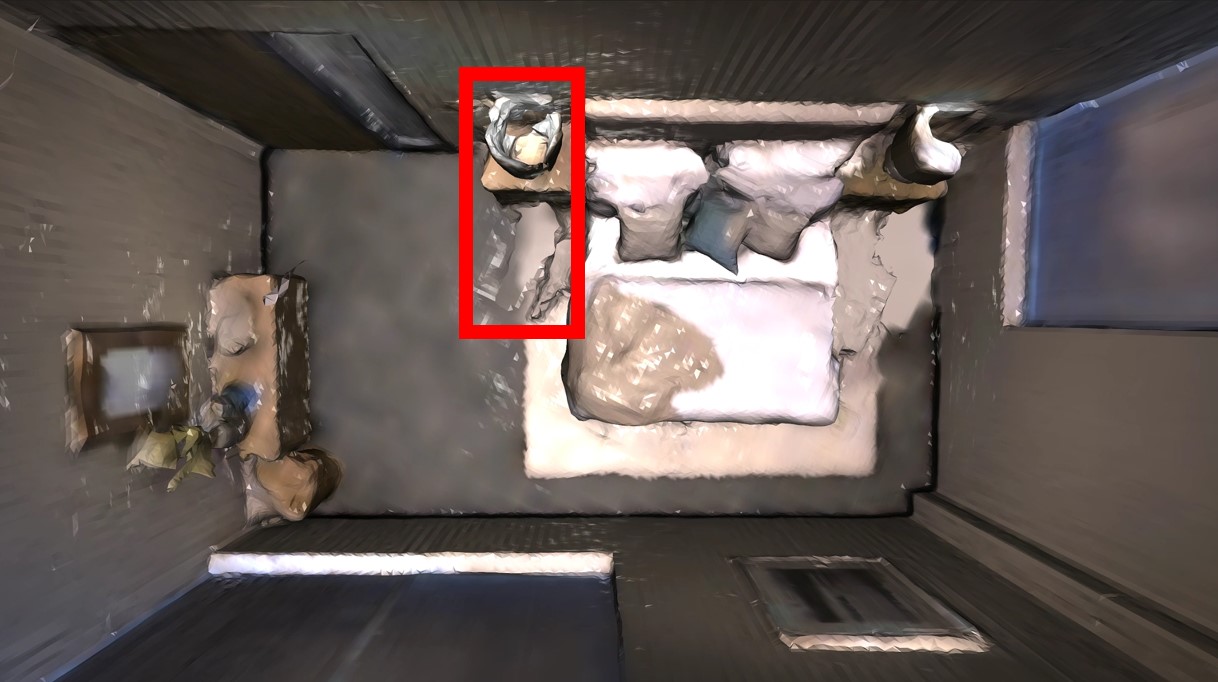,width=2.6cm}}
	\end{minipage}
	\vspace{1pt}
	\begin{minipage}[c]{0.32\linewidth}
		\centering
		{\epsfig{figure=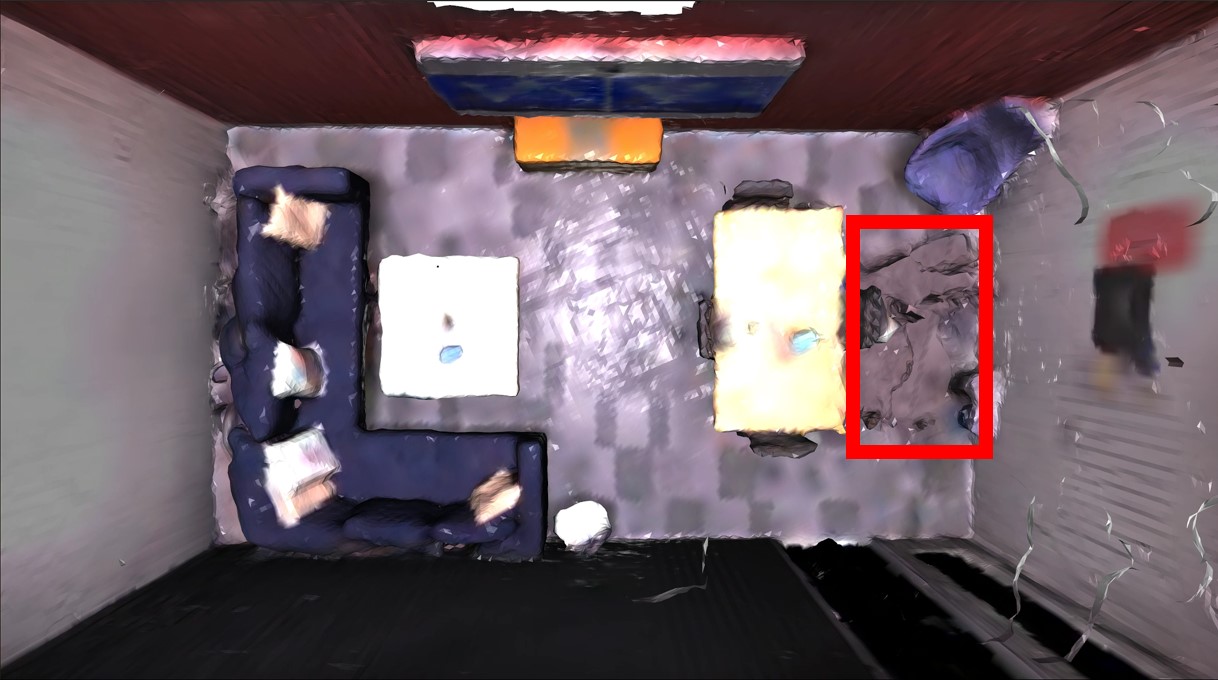,width=2.6cm}}
	\end{minipage}
	\vspace{1pt}
	\centering
	\begin{minipage}[c]{0.01\linewidth}
		\centering
		\rotatebox{90}{\tiny\texttt{GT}}
	\end{minipage}%
	\begin{minipage}[c]{.32\linewidth}
		\centering
		{\epsfig{figure=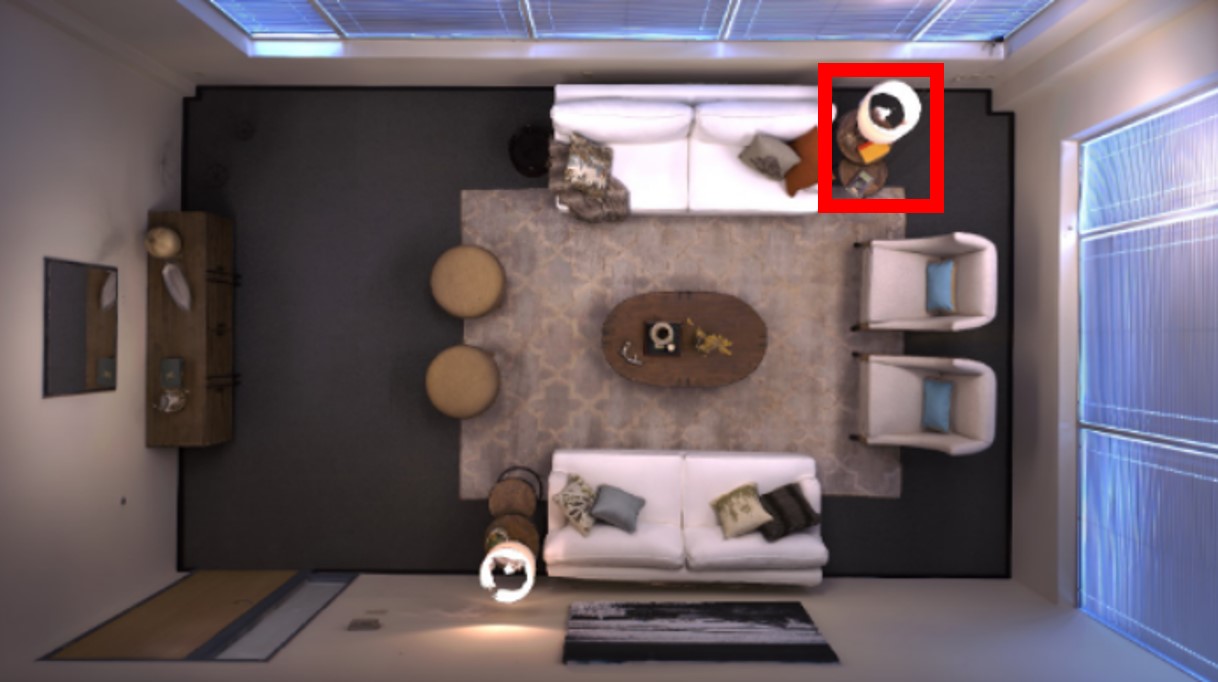,width=2.6cm}}
	\end{minipage}
	\begin{minipage}[c]{0.32\linewidth}
		\centering
		{\epsfig{figure=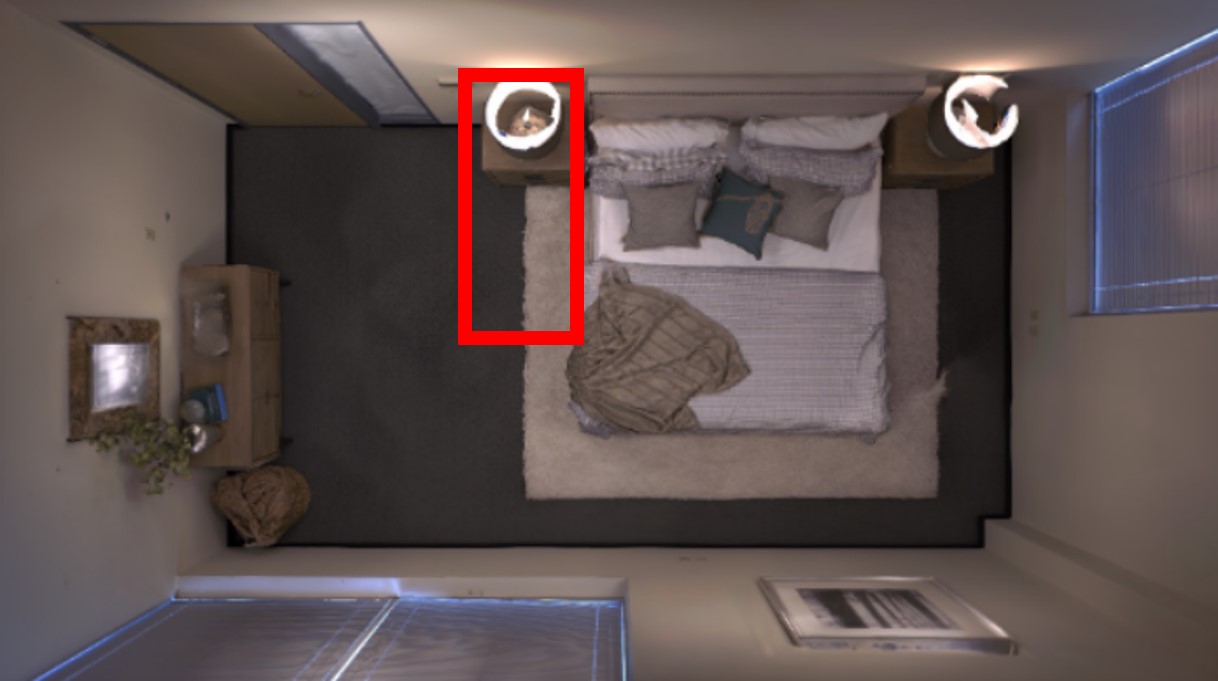,width=2.6cm}}
	\end{minipage}
	\begin{minipage}[c]{0.32\linewidth}
		\centering
		{\epsfig{figure=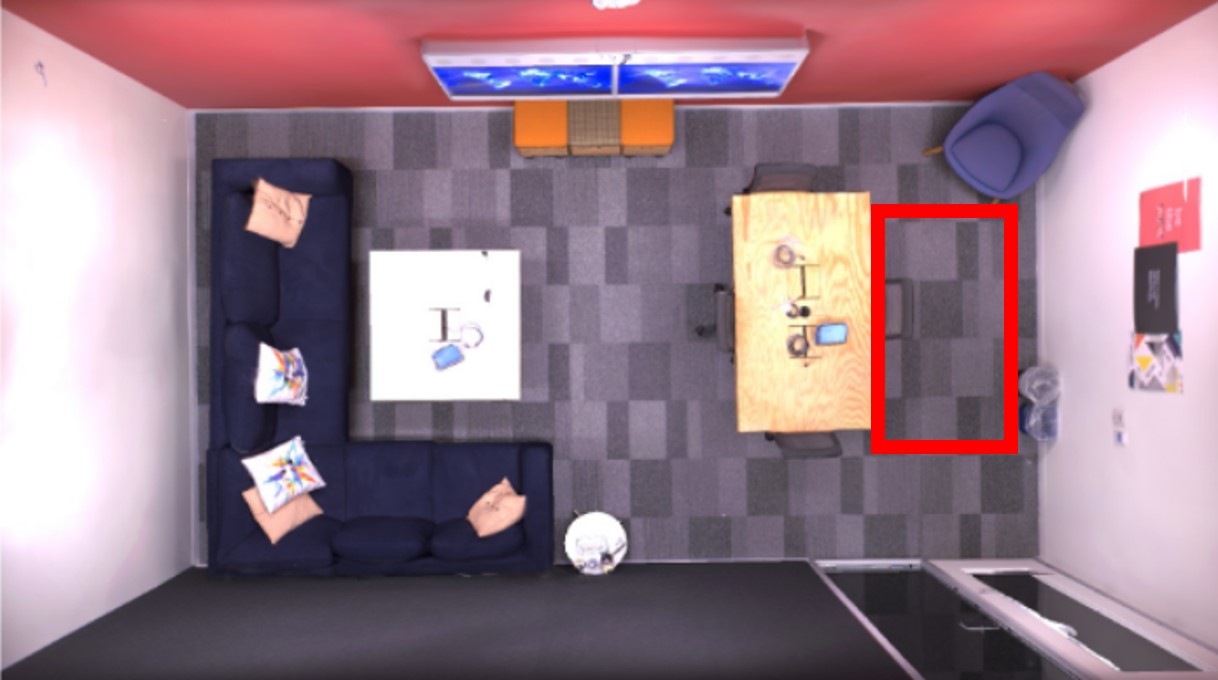,width=2.6cm}}
	\end{minipage}
	\caption{Reconstruction results on Replica dataset. The red box highlights the improved areas.} 
	\label{fig:Figure4}
\end{figure} 

\begin{table}[t!]
	\begin{center}
		\small
		\caption{Ablation study of our design choices on the TUM sequence fr3/w/half and fr3/w/xyz.}\label{tab:Table 4.}
		\resizebox{\linewidth}{!}{
			\begin{tabular}{c|c|c|cc}
				\hline
				Depth & Keyframe & Sampling & \multirow{2}{*}{fr3/w/half} & \multirow{2}{*}{fr3/w/xyz}\\
				Revision & Strategy & Strategy & & \\
				\hline
				$\times$ & $\checkmark$ & $\checkmark$ & 0.094 & 0.069\\
				$\checkmark$ & $\times$ & $\checkmark$ & 0.099 & 0.073\\
				$\checkmark$ & $\checkmark$ & $\times$ & 0.094 & 0.067\\
				$\checkmark$ & $\checkmark$ & $\checkmark$ & \textbf{0.071} & \textbf{0.064}\\
				\hline
			\end{tabular}
		}
	\end{center}
\end{table}

\subsection{Ablation study}

\noindent\textbf{Depth revision. }We demonstrate the effectiveness of depth revision on TUM \cite{15}. As shown in Table \ref{tab:Table 4.}, failing to revise depth information and further refining semantic masks not only significantly reduces tracking accuracy but also leads to less robustness.

\noindent\textbf{Keyframe selection. }We test our method using overlap-based strategy (as shown in the second row of Table \ref{tab:Table 4.}), where they simply perform equidistant selection. In contrast, our specially designed strategy for dynamic scenes not only enhances tracking robustness but also enriches the keyframe set with extensive scene information, playing a vital role in accurate tracking.

\noindent\textbf{Sampling. }In Table \ref{tab:Table 4.}, we also illustrate the effectiveness of our pixel sampling strategy. Results indicate that performing surface-based sampling and filtering out irrelevant points for rendering enhances the quality of sampled points, thereby improving the accuracy of camera tracking.

\section{Conclusion}
We introduce NID-SLAM, a dynamic RGB-D neural SLAM approach. We demonstrate that neural SLAM is capable of achieving high-quality mapping and plausible hole filling in dynamic scenes. Leveraging dynamic objects removal, our approach enables stable camera tracking and creates reusable static maps. The accurately obtained images, with dynamic objects removed, may also be utilized in further applications such as robotic navigation.

\noindent\textbf{Limitations. }The real-time performance of our method is currently limited by the speed of segmentation network. Finding a balance between segmentation quality and speed remains a direction for future research. Instead of solely relying on the static map, integrating the predictive capabilities of neural networks could lead to more comprehensive and accurate background inpainting.

\section*{Acknowledgment}
This work was partially supported by the Key R{\&}D Program of Zhejiang Province (No. 2023C01181)

\bibliographystyle{IEEEtran}
\bibliography{NID-SLAM-final-paper}
\end{document}